\documentclass[10pt,twocolumn,letterpaper,dvipsnames]{article}

\usepackage{cvpr}      % To produce the REVIEW version
\usepackage{graphicx}
\usepackage{amsmath}

\usepackage{amssymb}
\usepackage{booktabs}
\usepackage{multirow}
\usepackage{multicol}
\usepackage[table]{xcolor}% http://ctan.org/pkg/xcolor

\usepackage{dcolumn}

\usepackage{array}

\usepackage{pifont}% http://ctan.org/pkg/pifont
\definecolor{limegreen}{HTML}{32CD32}
\newcommand{\cmark}{\textcolor{limegreen}{\ding{51}}}%
\newcommand{\xmark}{\textcolor{red}{\ding{55}}}%

\usepackage[pagebackref,breaklinks,colorlinks]{hyperref}

\usepackage[capitalize]{cleveref}
\crefname{section}{Sec.}{Secs.}
\Crefname{section}{Section}{Sections}
\Crefname{table}{Table}{Tables}
\crefname{table}{Tab.}{Tabs.}

\def\cvprPaperID{1135} % *** Enter the CVPR Paper ID here
\def\confName{CVPR}
\def\confYear{2023}

\begin{document}

\title{DyNCA: Real-time Dynamic Texture Synthesis \\ Using Neural Cellular Automata}

\author{Ehsan Pajouheshgar$^*$, Yitao Xu$^*$, Tong Zhang, Sabine Süsstrunk\\
School of Computer and Communication Sciences, EPFL, Switzerland \\
{\tt\small \{ ehsan.pajouheshgar, yitao.xu, tong.zhang, sabine.susstrunk \}@epfl.ch}
% \and
\vspace{-12pt}
}

\newcolumntype{G}{>{\centering\arraybackslash} m{75pt} }  %# New column type
\newcolumntype{F}{>{\centering\arraybackslash} m{80pt} }  %# New column type

\newcommand{\arrow}{\textcolor{black}{\Longrightarrow}}%
\newcommand{\ofarrow}{\hspace*{-8pt}$ \overset{OF}{\arrow}$}

\twocolumn[{%
\renewcommand\twocolumn[1][]{#1}%
\maketitle
    \centering
    \captionsetup{type=figure}
    \includegraphics[width=\linewidth,trim={0cm 0 0 3pt},clip]{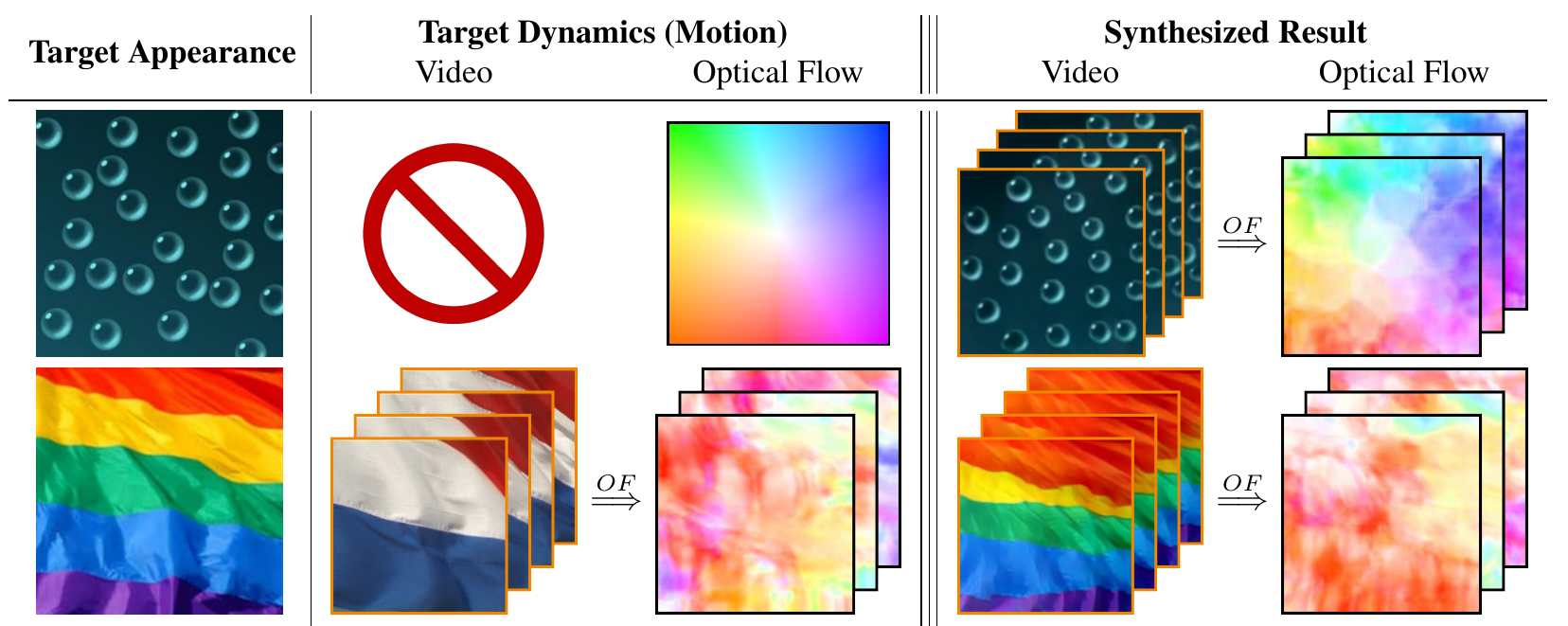}
    % \vspace{50pt}
    \vspace{-19pt}
    \captionof{figure}{ 
    Our DyNCA model can synthesize \textit{\textbf{infinitely-long}} realistic dynamic texture videos with \textit{\textbf{arbitrary size}} in \textit{\textbf{real time}}.
    \textbf{Target Appearance:} DyNCA learns a desired texture pattern from a given target appearance image.
    \textbf{Target Dynamics:} DyNCA can learn motion from different target sources. We allow the users to define the desired motion either by a hand-crafted optical-flow image\protect\footnotemark or a dynamic texture video. \textbf{Synthesized Result:} DyNCA synthesizes realistic dynamic texture videos. Each synthesized video frame resembles the target appearance, while the concatenation of frames induces the motion of the target dynamics. See our real-time interactive demo at\protect\footnotemark.}
    \vspace{12pt}
    \label{fig:teaser}
}]

% \maketitle

% \caption{\TZ{HAHAHA the first time that we are using more gpu than vilab  haha}}

% \maketitle

% \usepackage{multirow}

% \input{figures/teaser}

%%%%%%%%% ABSTRACT
\begin{abstract}
    \vspace{-10pt}
    Current \textbf{Dy}namic \textbf{T}exture \textbf{S}ynthesis (\textbf{DyTS}) models can synthesize realistic videos. However, they require a slow iterative optimization process to synthesize a single fixed-size short video, and they do not offer any post-training control over the synthesis process. We propose \textbf{Dy}namic \textbf{N}eural \textbf{C}ellular \textbf{A}utomata (DyNCA), a framework for real-time and controllable dynamic texture synthesis. Our method is built upon the recently introduced NCA models and can synthesize infinitely long and arbitrary-sized realistic video textures in real time. We quantitatively and qualitatively evaluate our model and show that our synthesized videos appear more realistic than the existing results. We improve the SOTA DyTS performance by $2\sim 4$ orders of magnitude.  Moreover, our model offers several real-time video controls including motion speed, motion direction, and an editing brush tool. We exhibit our trained models in an online interactive demo that runs on local hardware and is accessible on personal computers and smartphones.
\end{abstract}

%%%%%%%%% BODY TEXT
\vspace{-25pt}
\section{Introduction}
\vspace{-2pt}
\label{sec:intro}
Textures are everywhere. We perceive them as spatially repetitive patterns. \textit{Dynamic Textures} are textures that change over time and induce a sense of motion. Flames, sea waves, and fluttering branches are everyday examples.  Understanding and computationally modeling dynamic textures is an intriguing problem, as these patterns are observed in most natural scenes.
%We perceive them as  repetitive patterns in space (e.g., grass), or in time (e.g., pendulum).

\addtocounter{footnote}{-2} %3=n
 \stepcounter{footnote}\footnotetext{We use the same flow visualization as Baker et al. \cite{flow_visualization}.}
\stepcounter{footnote}\footnotetext{Link to the demo: \href{https://dynca.github.io}{https://dynca.github.io}}

%. Understanding and computationally modeling these patterns is an intriguing problem.

The goal of \textbf{D}ynamic \textbf{T}exture \textbf{S}ynthesis (\textbf{DyTS}) \cite{doretto2004spatially,doretto2003dynamic-firstds, soatto2001dynamic, ghadekar2014nonlinear,Holynski_2021_CVPR, costantini2007higherordersvd, funke2017synthesising-gatysdynamic, two_stream, zhang2021dynamic, xie2017generativeconvnet, yang2016stationary} is to generate perceptually-equivalent samples of an exemplar video texture\footnote{We use "video texture" and "dynamic texture" interchangeably.}. Applications of DyTS include the creation of special effects for backdrops and video games \cite{schodl2000video}, dynamic style transfer \cite{two_stream}, and creating cinemagraphs \cite{Holynski_2021_CVPR}.
% \ep{Why dynamic texture synthesis is important and that we focus on dynamic texture synthesis}
% Maybe we can say it has a lot of applications. (is it necessary?)

The state-of-the-art (SOTA) dynamic texture synthesis methods \cite{two_stream, xie2017generativeconvnet, zhang2021dynamic, funke2017synthesising-gatysdynamic, yang2016stationary} follow the same trend. They aim to find better optimization strategies to iteratively update a randomly initialized video until the synthesized video resembles the exemplar dynamic texture. Although the SOTA methods are able to synthesize acceptable quality videos, their optimization process is very slow and requires several hours to generate a \emph{single} \emph{fixed-resolution} short video %with 12 frames 
on a high-end GPU. Moreover, these methods do not offer any post-training control over the synthesized video.

\begin{figure}[t] 
	\centering
	\includegraphics[width=\linewidth, trim={0cm 11.42cm 14.25cm 0cm},clip]{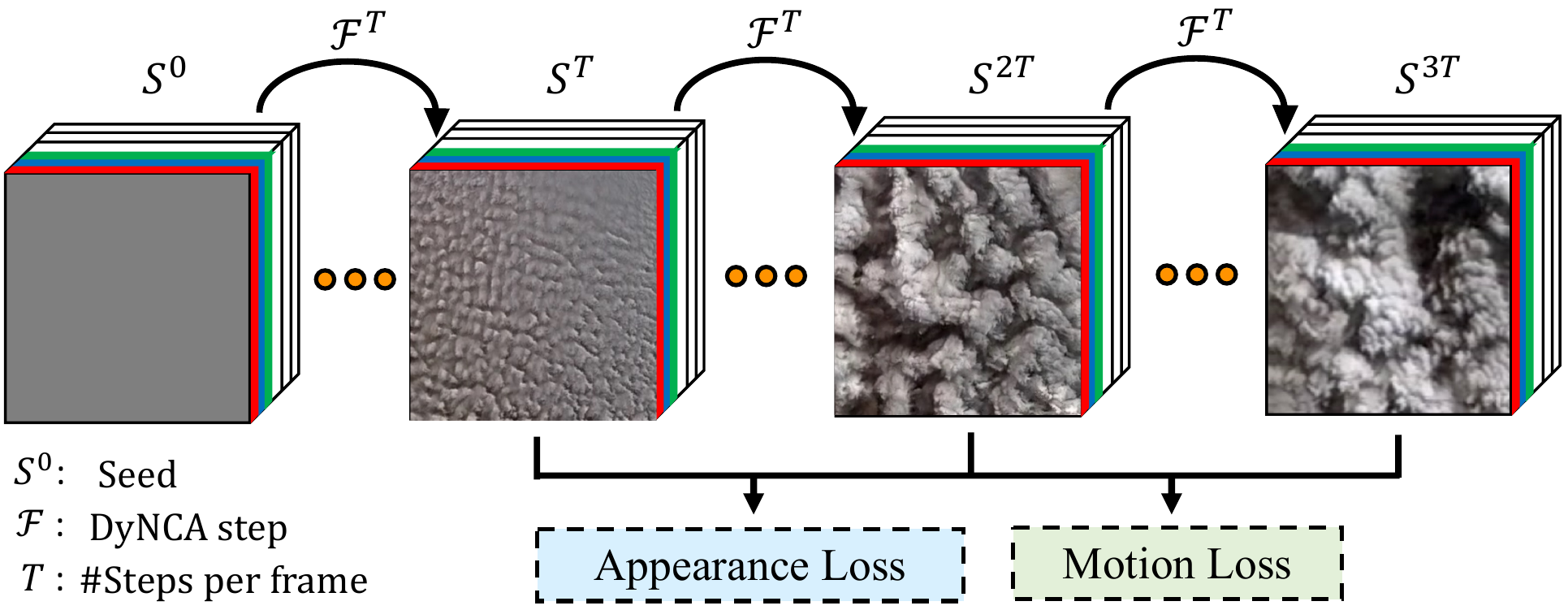}
	\caption{
	Overview of DyNCA. Starting from a seed state, DyNCA iteratively updates it, generating an image sequence. We extract images from this sequence and compare them with an appearance target as well as a motion target to obtain the DyNCA training objectives. After training, DyNCA can adapt to seeds of different heights and widths, and synthesize videos with arbitrary lengths. Sequentially applying DyNCA updates on the seed synthesizes dynamic texture videos in real time. 
% 	DyNCA starts from seed state, iteratively updating it to synthesize a video. The synthesized video is then compared with an appearance target, and a motion target to provide the loss functions for training the DyNCA model.
    }
	\label{fig:pipeline}
	\vspace{-15pt}
\end{figure}

In this paper we propose \textbf{Dy}namic \textbf{N}eural \textbf{C}ellular \textbf{A}utomata (\textbf{DyNCA}), a model for dynamic texture synthesis that is \emph{fast to train}, and once trained, can synthesize
%infinitely-long 
\emph{infinitely-long}, \emph{arbitrary-resolution} dynamic texture videos in \emph{real time} on a low-end GPU. Moreover, our method enables several real-time video editing controls, including motion direction and motion speed. Through quantitative and qualitative analysis we demonstrate that our synthesized videos achieve better quality and realism than the previous results in the literature.

Our model builds upon the recently introduced \textbf{N}eural \textbf{C}ellular \textbf{A}utomata (NCA) model \cite{mordvintsev2020growing, niklasson2021self-sothtml}. While Niklasson et al. \cite{niklasson2021self-sothtml} train the NCA with the goal of synthesizing static textures only, the NCA model is able to spontaneously generate randomly moving patterns. As an inherent property of NCA, these spontaneous motions are, however, unstructured and uncontrolled. We modify the architecture and the training scheme of NCA so that it can learn to synthesize video textures that have the desired motion and appearance. 

In short, our DyNCA model acts as a stochastic \textbf{P}artial \textbf{D}ifferential \textbf{E}quation (PDE), parameterized by a small neural network. DyNCA starts from a constant initial state called seed, and then iteratively evolves the state according to its trainable PDE update rule to generate a sequence of images. This image sequence is then evaluated against the appearance exemplar and the motion target to calculate the loss functions for the optimization of DyNCA, as illustrated in Figure~\ref{fig:pipeline}. We allow the user to specify the desired motion either by a motion vector field or an exemplar dynamic texture video. Moreover, by using a different target for the appearance and the motion, our model can perform dynamic style transfer, as shown in Figure~\ref{fig:teaser}. Our contributions summarized are:
\begin{itemize}
    \item Our DyNCA model, once trained, can synthesize dynamic texture videos in real time. 
    
    \item Our synthesized videos are on-par with or even better than the existing results in terms of realism and quality.
    
    \item After training, our DyNCA model can synthesize infinitely-long videos with arbitrary frame sizes.
    
    \item Our DyNCA framework enables several real-time interactive video editing controls including speed, direction, a brush tool, and local coordinate transformation. 
    
    \item We can perform real-time dynamic style transfer by learning appearance and motion from distinct sources.
\end{itemize}

\section{Related Works}
\label{sec:related_works}
% \vspace{-5pt}
In the following section, we discuss the existing DyTS methods. For comparison, Table~\ref{tab:method_comparison} summarizes the strengths (\cmark) and shortcomings (\xmark) of these methods.

% \vspace{-5pt}
\subsection{Dynamic Texture Synthesis}
% \vspace{-5pt}
In the literature, there are two dominant techniques to synthesize dynamic texture videos. \textbf{Recent approaches} follow the seminal work of Gatys et al.\cite{gatys2015texture} for texture synthesis. The authors propose an optimization-based method that relies on the features extracted by a deep neural network trained for image classification. Gatys et al.\cite{gatys2015texture} show that the Gram matrices of features extracted by the VGG network \cite{vgg} capture the perceptual qualities of texture images. Funke et al.\cite{funke2017synthesising-gatysdynamic} extend Gatys's idea to dynamic texture synthesis, and use a cross-frame Gram matrix of VGG features to capture the temporal characteristics of dynamic textures.
% in which the video textures are modeled by the statistics of concatenated feature maps through a certain time window.
Tesfaldet et al.\cite{two_stream} and Zhang et al. \cite{zhang2021dynamic} use a pre-trained optical flow network for extracting the motion features, which allows them to disentangle the appearance and motion aspects of video textures. Xie et al.\cite{xie2017generativeconvnet} 
%uses the idea of generative modeling to synthesize dynamic texture. They 
use an energy-based model characterized by a spatio-temporal 3D ConvNet and synthesize textures by sampling using Langevin dynamics. %Holynski et al.\TZ{what's wrong here?} %\cite{Holynski_2021_CVPR} adapt a different approach. They animate textures by moving the pixels according to a motion vector field, and use inpainting to fill the holes. Like DyNCA, they can also synthesize images with arbitrary sizes without retraining. However, their method is limited to the domain of textures observed during training.   

% \ep{should we mention artefacts and overfitting here?}
These methods utilize the expressivity of neural networks to produce realistic high-quality dynamic texture videos. However, they require a long training time to synthesize a single short video. Moreover, none of these methods provide any post-training controls over motion speed, frame size, and motion direction. These shortcomings make these methods unsuitable for real-time applications.

\textbf{Earlier DyTS methods} that can potentially enable real-time synthesis utilized PDEs to model dynamic textures \cite{doretto2003dynamic-firstds, costantini2007higherordersvd, yuan2004synthesizing-clds, ghadekar2014nonlinear, you2016kernel}. To synthesize video textures, Doretto et al.\cite{doretto2003dynamic-firstds} propose to use a linear dynamical system (LDS) in which each frame of the video is controlled by a latent variable whose evolution through time is driven by random noise. The
latent variable is obtained via projection of the target texture
image into a lower dimensional space by singular value decomposition (SVD). Costantini et al. \cite{costantini2007higherordersvd} propose to use higher-order SVD to improve the expressivity of LDS-based models. Yuan et al.\cite{yuan2004synthesizing-clds} introduce feedback into the LDS to improve the stability of the dynamical system. All these methods can synthesize new video textures without re-training, and potentially in real time. However, the synthesized videos are of low quality and contain artifacts.%\TZ{can we say something about the intuition why the have artefacts?}\yx{a very general explanation is lack of model capacity, or strong prior imposition on the representation of the video (Maybe SVD result contains just part of information)}. 

% \usepackage{graphicx}
\newcommand\RotText[1]{\rotatebox{90}{\parbox{2cm}{\centering#1}}}

\setcounter{footnote}{0}

\begin{table}[t]
\resizebox{\linewidth}{!}{%
\begin{tabular}{cccccccc}
\toprule
% \begin{tabular}[c]{@{}c@{}}\textbf{Sketch Synthesis} \\ \textbf{Algorithms}\end{tabular} & 
\begin{tabular}[c]{@{}c@{}}{Method}\end{tabular} &
\begin{tabular}[c]{@{}c@{}}{A}\end{tabular} &
\begin{tabular}[c]{@{}c@{}}{B}\end{tabular} 
&\begin{tabular}[c]{@{}c@{}}{C}\end{tabular} & \begin{tabular}[c]{@{}c@{}}{D}\end{tabular} & \begin{tabular}[c]{@{}c@{}}{E}\end{tabular} & \begin{tabular}[c]{@{}c@{}}{F}\end{tabular} & \begin{tabular}[c]{@{}c@{}}{G}\end{tabular} %& \begin{tabular}[c]{@{}c@{}}{H}\end{tabular}
\\
% \begin{tabular}[c]{@{}c@{}}{\textbf{G}}\end{tabular} & \\

\midrule 

Costantini et al. \cite{costantini2007higherordersvd} &\xmark & \cmark & \cmark & \xmark & \cmark & \xmark  & \xmark %& \cmark
\\

Doretto et al. \cite{doretto2003dynamic-firstds} &\xmark & \cmark & \cmark & \xmark & \cmark & \xmark  & \xmark %& \cmark
\\

Funke et al. \cite{funke2017synthesising-gatysdynamic} & \xmark & \cmark & \xmark & \xmark & \xmark & \xmark & \xmark %& \cmark
\\

Tesfaldet et al. \cite{two_stream} &\xmark & \cmark & \xmark & \xmark & \xmark & \cmark & \cmark %& \cmark
\\

Xie et al. \cite{xie2017generativeconvnet} &\xmark & \xmark & \xmark & \xmark & \cmark & \xmark & \xmark %& \cmark
\\

Zhang et al. \cite{zhang2021dynamic} &\xmark & \cmark & \xmark & \xmark & \xmark & \cmark  & \cmark %& \cmark
\\

% Holynski et al. \cite{Holynski_2021_CVPR} &\cmark & \xmark\footnotemark & \xmark & \xmark & \cmark & \cmark & \cmark %& \xmark
% \\

\midrule
\textbf{DyNCA (Ours)} & \cmark & \cmark & \cmark & \cmark & \xmark & \cmark &  \cmark %& \cmark
\\
\bottomrule
\end{tabular}
}
\vspace{-8pt}
\caption{Comparison of DyTS methods. \textbf{(A)} Can synthesize videos with arbitrary frame size after training. \textbf{(B)} Can synthesize arbitrarily long videos. \textbf{(C)} Can synthesize new video samples without re-training. \textbf{(D)} Allows real-time video editing (speed, direction, and a brush tool). \textbf{(E)} Does not rely on pre-trained models to extract motion or texture information. \textbf{(F)} Has disentangled appearance and motion. %\textbf{(G)} High-quality colored frames,
\textbf{(G)} Can learn motion from vector fields. %\textbf{(H)} Can synthesize time-varying motion.
}
\vspace{-13pt}
\label{tab:method_comparison}
\end{table}
% \footnotetext{Holynski et al. \cite{Holynski_2021_CVPR} synthesize video loops that are perceived as endless.}

% Comparison of GAN-based image editing algorithms by their characteristics. \fbox{\textbf{A}} Works on any Dataset, \fbox{\textbf{B}} No test-time optimization, \fbox{\textbf{C}} Works on any GAN architecture, \fbox{\textbf{D}} Can perform the edit using a single image, \fbox{\textbf{E}} Allows global semantic editing, \fbox{\textbf{F}} Allows localized semantic editing, \fbox{\textbf{G}} Allows editing any object in the image

Our DyNCA method benefits from the best of both approaches by combining the expressivity of neural networks with PDEs. Having $2\sim4$ orders of magnitude fewer parameters than the SOTA models, our model can synthesize realistic dynamic texture videos in real time.  %Next, we review the NCA model which constitutes the backbone of our DyNCA. 

% Because of utilizing PDEs instead of raw pixel values to characterize texture videos, our DyNCA have $2\sim4$ orders of magnitude less trainable parameters than the SOTA DyTS methods. 

% In contrast to SOTA DyTS methods which 

% We build upon the recently proposed NCA model \cite{mordvintsev2020growing, niklasson2021self-sothtml} and propose our DyNCA model for dynamic texture synthesis.\TZ{we don not have to re-state what we are doing, just summarize the reason will be fine. Actually the comment one is better.}

% it is hard for them to control the generated dynamic texture (length extension, speed control, etc.) in real-time after training.  Moreover, those methods typically optimize in pixel space with a large number of parameters to train. Instead, our method utilizes a model with a very little amount of parameters and adopts a temporal update scheme to generate complex patterns.

% \vspace{-5pt}
\subsection{NCA for Texture Synthesis}
% \vspace{-1pt}

% \ep{explain why NCA is a good model for texture synthesis and motion}
Gilpin \cite{gilpin2019cellular} shows that Cellular Automata models can be represented using Convolutional Neural Networks. Extending \cite{gilpin2019cellular}, Mordvinstev et al. \cite{mordvintsev2020growing} propose the Neural Cellular Automata model and show its potential in creating various self-organizing systems \cite{mordvintsev2020growing, mordvintsev2021mu-micronca, randazzo2020self-classifying, niklasson2021self-sothtml}. NCA is inspired by Turing's seminal work \cite{turing1990chemical} on pattern generation, and the observation that many natural patterns stem from local interactions between tiny particles, cells, or molecules \cite{niklasson2021self-sothtml}. 

% \sabine{Do take into account Tong's comments and add some stuff (1-2 sentences) here about NCA's history and applications. Then in a new paragraph... If at the end you are running out of space, you can move anything additional to supplementary material}

Niklasson et al.\cite{niklasson2021self-sothtml, mordvintsev2021mu-micronca} were the first to train NCA models to synthesize textures. 
While the NCA training signal originates from a static exemplar texture image, the model is able to spontaneously generate stable but randomly moving textures. We modify the architecture and training scheme of NCA to enable synthesizing structured motion. First, our DyNCA model  receives supervision from a pre-trained optical-flow network, which enables it to synthesize a video texture with structured motion. Moreover, we incorporate multi-scale perception and positional encoding into the DyNCA architecture. The proposed architectural changes increase the communication range of the cells and allow the cells to be aware of global information, respectively.

\vspace{-2pt}
\section{DyNCA Architecture}
\vspace{-2pt}
\label{sec:dynca_arc}
In the following sections, we first review the NCA model and then present the architecture of our DyNCA.

\vspace{-1pt}
\subsection{Neural Cellular Automata (NCA)}
\vspace{-1pt}

\begin{figure*}[t] 
	\centering
	\includegraphics[width=\linewidth,trim={0cm 6.6cm 5.6cm 0cm},clip]{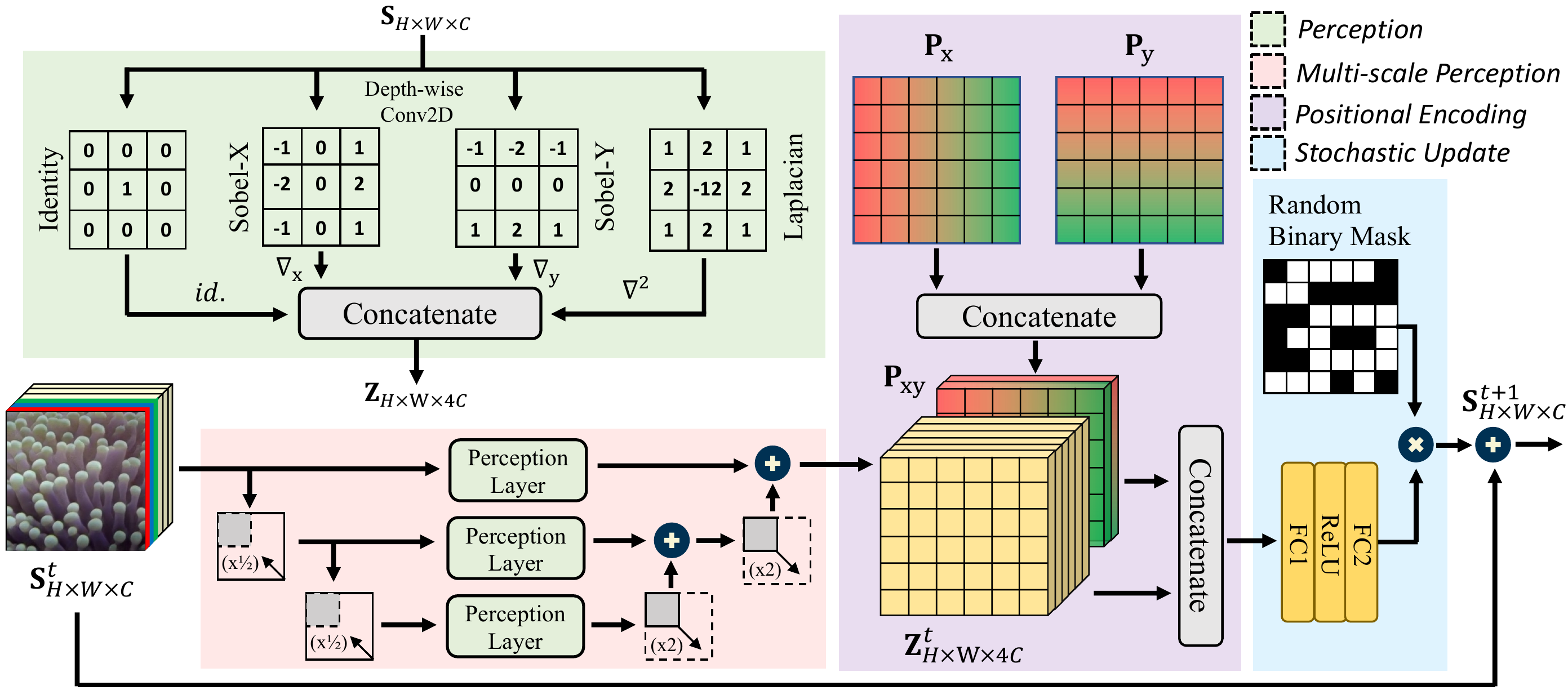}
	\vspace{-16pt}
	\caption{Illustration of a single DyNCA step. Given an input state $\mathbf{S}^t \in R^{H \times W \times C}$ at time step $t$, 
	each cell first perceives its neighbors on various scales with the same perception layer. 
	The perception tensor of each scale is then upsampled and summed up, and is concatenated with the positional encoding tensor $\mathbf{P}_{\textup{x}\textup{y}}$. 
	Each cell then applies the same update rule, parameterized by a small MLP. Finally, all cells perform a stochastic residual update to determine the state of the cells in time $t + \Delta t$. We use $\Delta t = 1$ in our model.
	}
	\label{fig:dynca_arch}
	\vspace{-12pt}
\end{figure*}

\label{sec:nca}
The idea of NCA stems from cellular automata, in which cells live on a grid, and each cell communicates with its neighbors to determine the cell's next state. In NCA, cell states at time $t$ are represented by $\mathbf{S}^{t} \in \mathbb{R}^{H \times W \times C}$ where $H \times W$ is the grid size. The $C$ dimensional vector $s^{t}_{ij}$ encodes the state of the cell at location $i, j$, where the first three dimensions define the RGB color of the cell.

\definecolor{perception_green}{HTML}{E5F2DE}
The NCA starts from a constant zero-filled initial state called seed
%\footnote{Similar to Niklasson et al. \cite{mordvintsev2021texture-sot} we use zeros constant as the seed}
and evolves this state over time according to its trainable PDE. The update rule of this PDE consists of two parts, \textit{Perception} and \textit{Stochastic Update}. At the perception stage, each cell gathers information from its surrounding neighbors, forming the perception vector $z_{ij} \in \mathbb{R}^{4C}$, illustrated in the \colorbox{perception_green}{green box} in Figure~\ref{fig:dynca_arch}. %where $i, j$ represent the index of the cell.
\vspace{-2pt}
\begin{equation}
    z_{ij} = Concat(s_{ij}, \nabla_{\textup{x}} \mathbf{S}|_{ij}, \nabla_{\textup{y}} \mathbf{S}|_{ij}, \nabla^2 \mathbf{S}|_{ij})
    \label{eq:perception}
\end{equation}
\vspace{0pt}
Note that the convolution kernels in the perception stage are frozen during training. In the stochastic update stage, the new state of each cell is determined based on its perception vector. 
The update stage of NCA can be viewed as a stochastic discrete-time, discrete-space PDE:

% \begin{equation}
%     S^{t + 1}_{ij} = \mathcal{F} (S^t) = S^t_{ij} + (MLP(Z_{ij}) \odot M)
%     \label{eq:nca_pde}
% \end{equation}
\vspace{-10pt}
\begin{equation}
\begin{split}
    \mathbf{S}&^{t + 1} = \mathcal{F} (\mathbf{S}^t) = \mathbf{S}^t + \frac{\partial \mathbf{S}}{\partial t} \Delta t\\
    & \frac{\partial s_{ij}}{\partial t} = \textup{MLP}(z_{ij}) \odot M
\end{split}
\label{eq:nca_pde}
\end{equation}
\vspace{-0pt}
\definecolor{update_blue}{HTML}{DFF3FE}
where $\textup{MLP}$ is a \textbf{M}ulti \textbf{L}ayered \textbf{P}erceptron with two layers and a ReLU activation function. The residual update values produced by the MLP are multiplied by a binary random variable $M$ to introduce stochasticity into the model. This ensures that the cells can work asynchronously, and also enables synthesizing new texture samples. The stochastic update stage is illustrated in the \colorbox{update_blue}{blue box} in Figure~\ref{fig:dynca_arch}.

To facilitate long-range cell communication and to allow the cells to be aware of global information, we introduce multi-scale perception and positional encoding into the vanilla NCA architecture, respectively.  Our experiments in section \ref{sec:exp} demonstrate their necessity for DyTS. The overall architecture of our DyNCA model is illustrated in Figure~\ref{fig:dynca_arch}. To the best of our knowledge, we are the first to incorporate these two architectural changes in conjunction with NCA models. In the following sections, we elaborate on the proposed architectural modifications.

\subsection{Multi-scale Perception}

\label{sec:multiscale-perception}

\definecolor{multi_scale_red}{HTML}{FFE8E7}
\vspace{-0pt}
In the perception stage of vanilla NCA, each cell only receives information from its eight surrounding cells. Therefore, many timesteps are needed for far-apart cells to perceive and communicate with each other. 
% If the predefined maximum NCA steps is not large enough, it can potentially cause incorrect pattern formation.
This problem becomes more pronounced when we increase the grid size $H\times W$, since the $3\times 3$ perception kernels become smaller in proportion to the image size. One straightforward solution is to increase the number of NCA steps during training to facilitate long-range cell communication. However, this simple solution increases memory usage, slows down the training, and makes the training more unstable. %\sabine{The following sentence does not go fit here, I put it below.}  

To solve this problem, we propose \textit{Multi-Scale Perception} (MSP). The idea of using multi-scale analysis predates deep learning and has shown its effectiveness in many computer vision tasks \cite{adelson1984pyramid, sift, burt1987laplacian, multiscale_turing}. We build a pyramid of cell states via bilinear downsampling and apply the same perception kernels at different scales. To combine the information from all scales, we upsample and sum up the perception vectors, as shown in the \colorbox{multi_scale_red}{red box} in Figure~\ref{fig:dynca_arch}.
%We can regard this procedure as performing dilated convolution in the perception stage.
Multi-scale perception increases the communication range of the cells and allows messages to pass between faraway cells in fewer steps. It also improves the stability of DyNCA and makes the training less sensitive to hyperparameters. We perform an ablation study of multi-scale perception in section \ref{sec:exp-ablation} and show its importance in preserving appearance fidelity.

% \vspace{-5pt}
\subsection{Positional Encoding}
\label{sec:positional-encoding}
% \vspace{-5pt}
\definecolor{positional_purple}{HTML}{E9DDF0}

In order to apply the $3\times3$ depth-wise convolutions in the perception stage, one needs to adopt a padding strategy to retain the spatial dimensions. Niklasson et al. \cite{niklasson2021self-sothtml} use circular padding to make the cells homogeneous and to reduce spatial inductive bias. While cell homogeneity is a good assumption for generating static textures, it is not suitable for synthesizing structured motion. Our intuition is that, in physical systems, motion not only arises from local interactions between tiny particles and cells, but also from global external forces such as gravity. Hence, we allow the cells to be aware of their position and propose to use replicate padding and \textit{Cartesian Positional Encoding} (CPE), which is known to provide a more consistent spatial inductive bias than zero-padding \cite{padding_inductive_bias}. As illustrated in the \colorbox{positional_purple}{purple box} in Figure~\ref{fig:dynca_arch}, we concatenate the output of the multi-scale perception stage with a two-channel tensor $\mathbf{P}_{\textup{xy}}$, where $\mathbf{P}_{ij} =\begin{bmatrix}
\frac{2i + 1.0}{W}
\\ 
\frac{2j + 1.0}{H}
\end{bmatrix} - 1.0$. Our ablation study in section~\ref{sec:exp-ablation} shows that employing CPE drastically improves motion consistency and accuracy in the synthesized videos. 

% \yx{do we need to mention replicate also good for local texture fitting} \ep{No because then we have to support this claim and explain more} 

% positional encoding (Cartesian)

% Problem with original NCA with guided motion: after resize the motion is not good (describe)

% Key: let the cell know the global position -> pos encoding (other method to be criticized?)

% compare those two:

% 1. pos encoding by transformer method: absolute position, problem: resize will cause unknown position encoding

% 2. by Cartesian: relative position, scale-persistent

% \vspace{-14pt}
\section{DyNCA Training}
\begin{figure}[t] 
	\centering
	\includegraphics[width=\linewidth, trim={0cm 7.4cm 3.6cm 0cm},clip]{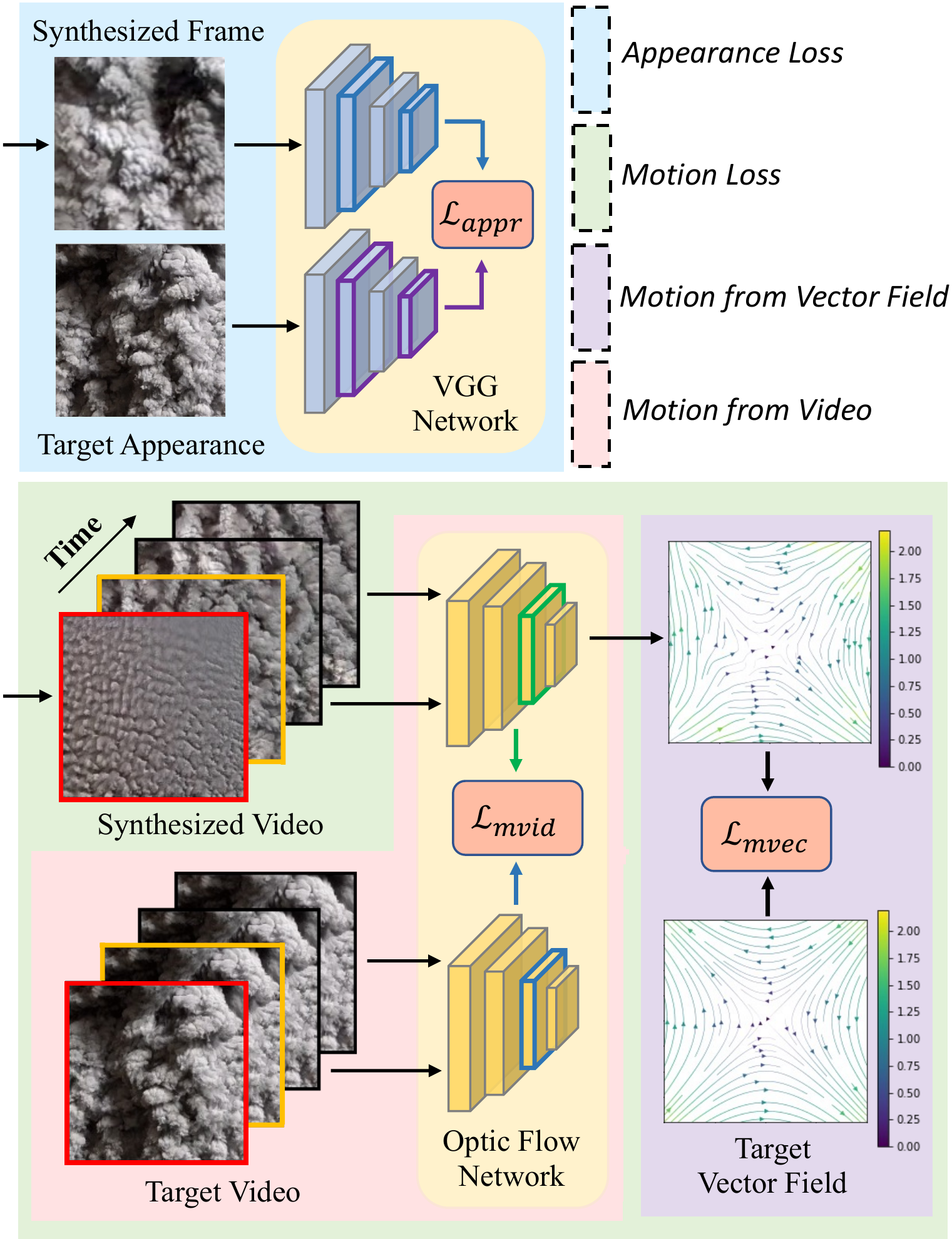}
	\vspace{-10pt}
	\caption{
	Overview of our loss functions. \textbf{Appearance loss ($\mathcal{L}_{appr}$)}: DyNCA learns the target appearance from a static image via minimizing an optimal-transport-based style matching loss \cite{kolkin2019style-otloss} between deep features extracted from the VGG16 network \cite{vgg}. \textbf{Motion Loss ($\mathcal{L}_{mvec}$)}: We guide DyNCA to synthesize a video having an optical flow similar to the target vector field. \textbf{Motion Loss ($\mathcal{L}_{mvid}$)}: DyNCA fits the target video motion by minimizing an optimal-transport-based style matching loss between deep features of a pre-trained optical-flow prediction network.}
	\label{fig:loss}
	\vspace{-15pt}
\end{figure}

% \vspace{-3pt}
Our DyNCA model acts as a PDE and generates a sequence of images $\mathcal{I}^g=\{\mathcal{I}^g_1, \mathcal{I}^g_2, ...\}$. We sample the synthesized video frames $\mathcal{V}^g=\{\mathcal{V}^g_1, \mathcal{V}^g_2, ...\}$ from this image sequence by mapping $T$ DyNCA steps to one frame, i.e. $\mathcal{V}^g_k = \mathcal{I}^g_{kT}$. The synthesized video is then evaluated against the target appearance and the target motion to compute the appearance loss $\mathcal{L}_{appr}$ and the motion loss $\mathcal{L}_{motion}$, respectively. This process is shown in Figure~\ref{fig:pipeline}. Our final loss is $\mathcal{L}_{\textup{DyNCA}} = \mathcal{L}_{appr} + \lambda \mathcal{L}_{motion}$. The overall scheme of the proposed loss functions is illustrated in Figure~\ref{fig:loss}. %In the following sections, we elaborate on the loss functions. 

% 	(1) \textbf{Appearance loss}: Given one static texture image as the target appearance, we feed it to VGG network and extract the feature maps in middle layers. We repeat this for a synthesized frame and compute the style matching loss\yx{based on OT?} between two groups of feature maps. (2) \textbf{Motion loss (vector field)}: For motion coming from vector fields, we first compute the optic flow images between consecutive frames in the synthesized and the target video. Then we change the point-wise motion strength and direction on the generated optic flow images according to the target optic flow images. (3) \textbf{Motion loss (video)}: For motion coming from videos, we also feed consecutive frames in the synthesized and the target video to the optic flow network but instead of directly looking at the optic flow images, we extract the intermediate features. We construct a style matching loss that has the same form as the appearance loss to match the style of motion in synthesized and target videos.

\vspace{-3pt}
\subsection{Appearance Loss}
\vspace{-3pt}
\label{sec:texture}

%The appearance loss is for DyNCA to learn a target static texture.
We follow the texture fitting scheme proposed by Niklasson et al.\cite{niklasson2021self-sothtml}, where the statistics of deep features of the NCA-generated images are forced to match the ones of the target texture. We use a pre-trained VGG16 network \cite{vgg} to extract the deep features from the images. We denote VGG as $\mathcal{F}_{VGG}$ and the feature map extracted from layer $l$ as $\mathcal{F}^{l}_{VGG}$. Niklasson et al. \cite{niklasson2021self-sothtml} use the Gram matrix to define the training objective of the NCA. However, using Gram matrices can cause unstable training\cite{risser2017stable-gramunstable}, and can result in textures with wrong colors \cite{mordvinstev_youtube, niklasson2021self-sothtml}. Hence, we use the style loss proposed by Kolkin et al.\cite{kolkin2019style-otloss}, which has shown to produce better results. This loss is composed of a structure-matching term and a moment-matching term. Given a feature map of size $C \times H \times W$, the algorithm first creates a set of deep features with $n = H \times W$ elements by flattening the feature map along the spatial dimensions. Let $X$ and $Y$ be the deep feature set of a synthesized image and target appearance image, respectively. Then the structure-matching term $\mathcal{L}_{s}$ and the moment-matching term $\mathcal{L}_{m}$ are defined as:
\vspace{0pt}
\begin{equation}
\begin{split}
    D(&A, B) =  \frac{1}{|A|} \sum_{i} \mathop{min}\limits_{j} \left ( 1-\frac{A_i \cdot B_j }{||A_i||_2  ||B_j||_2} \right ) \\
    & \mathcal{L}_{s}(X,Y) = max \left \{D(X, Y), D(Y, X) \right \},
\end{split}
\label{eq:ot-structure}
\end{equation}
\vspace{0pt}
\begin{equation}
\begin{aligned}
   \mathcal{L}_{m}(X,Y)=\frac{1}{C} \left \| \boldsymbol{\mu}_X - \boldsymbol{\mu}_Y \right \|_1 + \frac{1}{C^2} \left \| \boldsymbol{\Sigma}_X - \boldsymbol{\Sigma}_Y \right \|_1,
\end{aligned}
\label{eq:ot-moment}
\end{equation}
where $D$ measures the distance between two deep feature sets, and $\boldsymbol{\mu}$, $\boldsymbol{\Sigma}$ are the mean and covariance matrix of their corresponding set, respectively. Let $X^l_k$ and $Y^l$ be deep VGG features extracted by $\mathcal{F}^{conv\; l\_1}_{VGG}$, where $l$ and $k$ indicate the VGG block index and the synthesized image index, respectively. Our final appearance loss is then defined as:
\vspace{0pt}
\begin{equation}
    \mathcal{L}_{appr}=\frac{1}{K}\sum^{K}_{k=1} \sum^5_{l=1} \left (\mathcal{L}_{s} (X^l_k, Y^l) +  \mathcal{L}_{m} (X^l_k, Y^l)\right )
\end{equation}

% \sabine{you need to define D(A,B). Also not sure I would use X and Y...they are the same feature set. What about X subscript synI and X subscript tarI? Makes the following also easier to read and interpret. No strong opinion though, I can be convinced otherwise :-) }

% Let $\mathcal{L}^l_{ot}=\mathcal{L}^l_{s} + \mathcal{L}^l_{m}$ be the optimal-transport-based style matching loss base on $\mathcal{F}^l_{VGG}$ features. 

% We collect $\mathcal{L}^l_{ot}$ from a set of layers $l_{VGG}=\{{l_1,l_2,...}\}$ in the VGG network. There are five blocks in a VGG16 network, denoted as $block_{1-5}$. We select the first convolution-ReLU layer in each block and  build $l_{VGG}$ with length 5. We keep $l_{VGG}$ the same for both vector field motion and video motion learning. 

% Given a target static texture image $\mathcal{I}_t$, we first compare each synthesized frame $\mathcal{V}^g_k$ with $\mathcal{I}_t$ using $\mathcal{L}^k_{appr}=\sum_{l \in l_{VGG}}\mathcal{L}^l_{ot}$, where $k$ is the frame index of $\mathcal{V}^g$. Our final appearance loss is $\mathcal{L}_{appr}=\frac{1}{L}\sum^{L}_{k=1}\mathcal{L}^k_{appr}$.

% Our appearance loss is defined as $\mathcal{L}_{appr}=\sum_{l \in l_{VGG}}\mathcal{L}^l_{ot}$.

% Define a feature vector inside a feature map as one of its channels. The structure matching term is defined as a relaxed optimal transport distance between two set of feature vectors. \yx{formula}. The moment matching term is defined as the absolute difference between the mean and variance of the extracted feature map. 

\subsection{Motion Loss}
% \subsubsection{Optic Flow Prediction Network}
% \label{sec:ofnet}
\vspace{-2pt}
The motion loss guides the DyNCA to produce the desired motion. We use the pre-trained \textit{Optical-Flow Prediction Network} from \cite{two_stream} to quantify the motion information between two frames. We denote this network as $\mathcal{F}_{OF}$, and the feature map extracted from layer $l$ as $\mathcal{F}^{l}_{OF}$. For different types of target motion sources, namely vector fields and videos, our motion loss takes two different forms, $\mathcal{L}_{mvec}$ and $\mathcal{L}_{mvid}$, respectively. These terms are discussed below.

% For extracting motion information, we use MSOENet introduced in \cite{two_stream}. \yx{how to introduce this thing}

\vspace{-5pt}
\subsubsection{Motion from Vector Field}
% \vspace{-4pt}
\label{sec:mvec}

Let $U^t \in \mathbb{R}^{H \times W \times 2}$ be the target motion vector field. Let $U^g = \mathcal{F}_{OF} (\mathcal{I}^g_{t_1}, \mathcal{I}^g_{t_2})$ be the optical-flow prediction on two synthesized images, where $t_1, t_2$ are two random indices such that $t_2 > t_1$. We propose two losses, $\mathcal{L}_{dir}$ for matching the motion direction, defined as: 
\vspace{0pt}
\begin{equation}
\mathcal{L}_{dir} = \frac{1}{HW} \sum_{i, j} \left ( 1 - \frac{U^g_{ij} \cdot U^t_{ij}}{\left \| U^g_{ij} \right \|_2 \left \| U^t_{ij} \right \|_2 } \right ),
\label{eq:dir_loss}
\end{equation}
and $\mathcal{L}_{norm}$ for matching the motion magnitude, defined as:
\begin{equation}
\mathcal{L}_{norm} = \frac{1}{HW} \sum_{i, j}  \left | \frac{T}{t_2 - t_1}\left \| U^g_{ij} \right \|_2 - \left \| U^t_{ij} \right \|_2 \right |.
\label{eq:norm_loss}
\end{equation}
Since $t_1, t_2$ are selected randomly, we scale the norm of the $U^g$ by $\frac{T}{t_2 - t_1}$ before comparing it against the target motion vector field. We define our final loss $\mathcal{L}_{mvec}$ as:
\begin{equation}
\mathcal{L}_{mvec} = \left ( 1.0 - min\{1.0, \mathcal{L}_{dir}\} \right ) \mathcal{L}_{norm} + \gamma \mathcal{L}_{dir},
\label{eq:mvec_loss}
\end{equation} where $\gamma$ is a constant coefficient. We set the norm-matching loss weight to $\left (1.0 - min\{1.0, \mathcal{L}_{dir}\} \right )$ to guide the training process to first focus on producing motion with the correct direction before trying to match the norm of the synthesized motion with the target vector field.

\subsubsection{Motion from Video}
\label{sec:mvid}
\vspace{-3pt}
For training DyNCA to learn the motion from a target video, we aim to match the motion between successive frames of the synthesized and target videos. We synthesize a video $\mathcal{V}^g$ with $K$ frames, and randomly pick the same number of consecutive frames from the target video, forming $\mathcal{V}^t$. Inspired by Tesfaldet et al. \cite{two_stream}, we feed successive video frames to a pre-trained optical-flow network and extract deep features from the concatenation layer of the network, denoted as $\mathcal{F}^{concat}_{OF}$. To compare the deep optical flow features of the target and the synthesized videos, we use the same structure-matching and moment-matching terms defined in Section~\ref{sec:texture}. We denote the extracted deep optical-flow features as $X_k$ for the synthesized video, and $Y_k$ for the target video, where $k$ indicates the frame index. Notice that there are $K - 1$ successive frame pairs in total. We define the objective for learning motion from video textures as:
\vspace{-4pt}
\begin{equation}
\begin{aligned}
   \mathcal{L}_{mvid}=\frac{1}{K-1}
   \sum^{K-1}_{k=1} \left ( \mathcal{L}_{s}(X_k, Y_k) + \mathcal{L}_{m}(X_k, Y_k) \right )
\end{aligned}
\label{eq:ot-motiontexture}
\end{equation}
\vspace{-4pt}

\vspace{-12pt}
\section{Experiments}
\vspace{-2pt}
\label{sec:exp}

% \usepackage{multirow}

\newcommand{\imgteaser}[1]{\includegraphics[height=75pt]{figures/teaser/#1}}
\newcommand{\imgmvid}[1]{\includegraphics[height=75pt]{figures/Experiments/MotionVid/#1}}
\newcommand{\imgteaserS}[1]{\includegraphics[height=55pt]{figures/teaser/#1}}
\newcommand{\imgteaserM}[1]{\includegraphics[height=68pt]{figures/teaser/#1}}

\newcolumntype{G}{>{\centering\arraybackslash} m{75pt} }  %# New column type
\newcolumntype{F}{>{\centering\arraybackslash} m{80pt} } 
\newcolumntype{H}{>{\centering\arraybackslash} m{5pt} }  %# New column type
%# New column type

% \newcommand{\arrow}{\textcolor{black}{\Longrightarrow}}%
% \newcommand{\ofarrow}{\hspace*{-8pt}$ \overset{OF}{\arrow}$}

\begin{table*}[]
\begin{tabular}{HF|Gm{0pt}G|||Gm{0pt}G}

    \multicolumn{2}{c|}{\multirow{2}{*}{\textbf{Target Appearance}}} & \multicolumn{3}{c|||}{\hspace{-8pt}\textbf{Target Dynamics (Motion)}} & \multicolumn{3}{c}{\hspace{-4pt}\textbf{Synthesized Result}} \\
    
    \vspace{18pt}

    &                                    & Video     &        & Optical Flow     & Video      &        & Optical Flow     \\
    \midrule
    
% (a) & \imgteaser{bubble_app.jpg}                                & \imgteaserS{none.png}      &       & \imgteaserM{target_flow.png}            & \imgteaser{bubble_synth.png}        & \ofarrow     & \imgteaser{bubble_synth_flow.png}            \\

% (b) & \imgteaser{bubble_app.jpg}                                & \imgteaserS{none.png}      &       & \imgteaserM{target_flow.png}            & \imgteaser{bubble_synth.png}        & \ofarrow     & \imgteaser{bubble_synth_flow.png}            \\

\multirow{2}{*}[21pt]{\rotatebox{90}{\parbox{5.5cm}{\hspace{30pt} \textbf{Dynamic Texture Synthesis}}}}   & \imgmvid{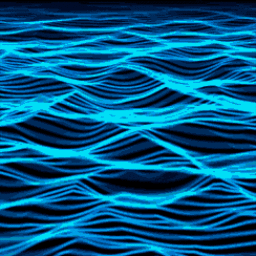}                                & \imgmvid{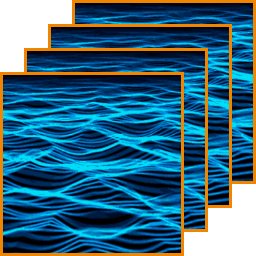}      &     \ofarrow   & \imgmvid{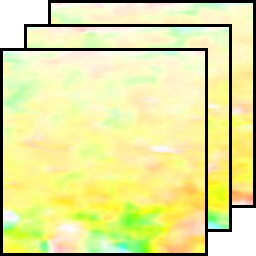}            & \imgmvid{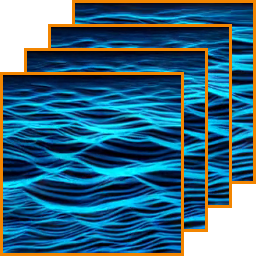}        & \ofarrow     & \imgmvid{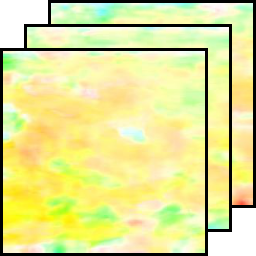}            \\

 & \imgmvid{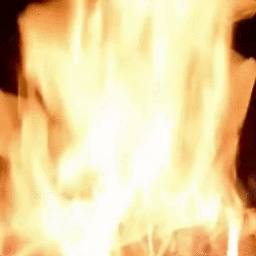}                                & \imgmvid{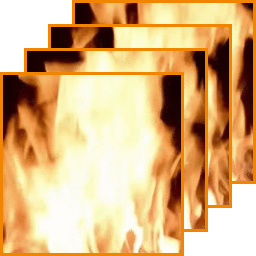}      &     \ofarrow   & \imgmvid{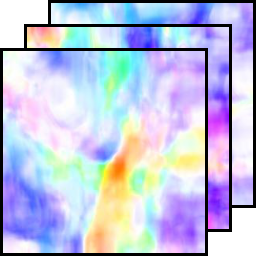}            & \imgmvid{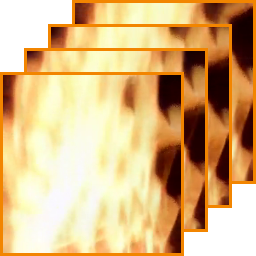}        & \ofarrow     & \imgmvid{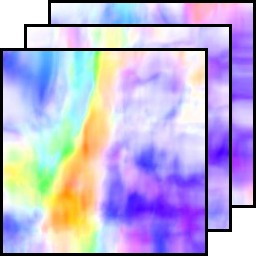}            \\

\midrule
\multirow{2}{*}[5pt]{\rotatebox{90}{\parbox{3.64cm}{\centering \textbf{\hspace{0pt} Dynamic Style Transfer}}}} & \imgmvid{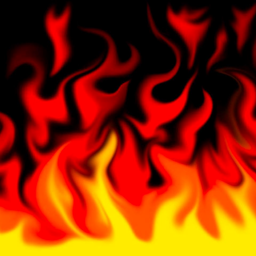}                                & \imgmvid{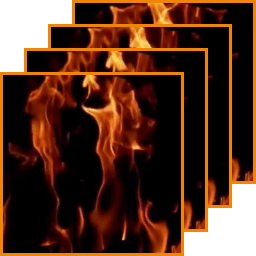}      &     \ofarrow   & \imgmvid{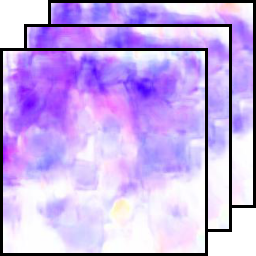}            & \imgmvid{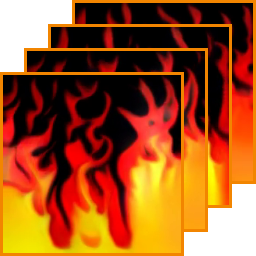}        & \ofarrow     & \imgmvid{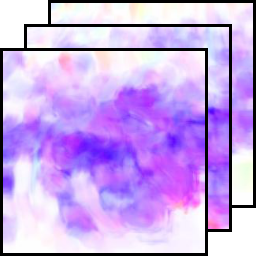}            \\

 & \imgmvid{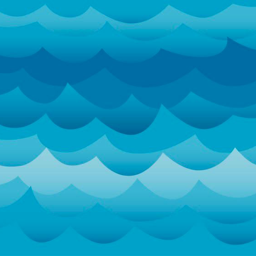}                                & \imgmvid{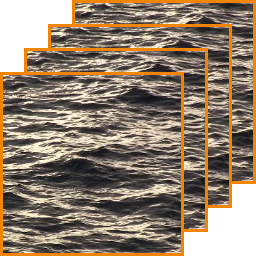}      &     \ofarrow   & \imgmvid{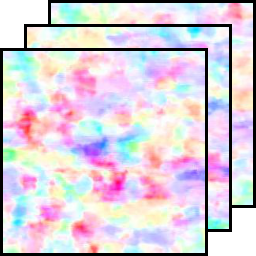}            & \imgmvid{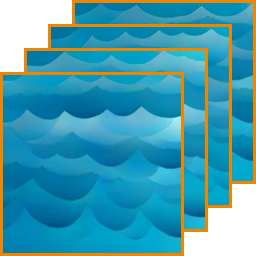}        & \ofarrow     & \imgmvid{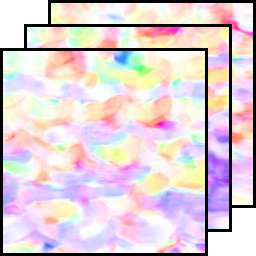}            \\

% (c) & \imgteaser{1.jpeg}                                & \imgteaser{1.jpeg}      &     \ofarrow   & \imgteaser{1.jpeg}            & \imgteaser{1.jpeg}        & \ofarrow     & \imgteaser{1.jpeg}            \\

% (c) & \imgteaser{1.jpeg}                                & \imgteaser{1.jpeg}      &     \ofarrow   & \imgteaser{1.jpeg}            & \imgteaser{1.jpeg}        & \ofarrow     & \imgteaser{1.jpeg}            \\

% (d) & \imgteaser{1.jpeg}                                & \imgteaser{1.jpeg}      &     ==   & \imgteaser{1.jpeg}            & \imgteaser{1.jpeg}        & ==     & \imgteaser{1.jpeg}            \\    
\end{tabular}
\vspace{-5pt}
\captionof{figure}{Results of dynamic texture synthesis and dynamic style transfer with DyNCA-L-$256^2$. \textbf{Dynamic Texture Synthesis}: DyNCA faithfully reproduces the target appearance and target dynamics. \textbf{Dynamic Style Transfer}: DyNCA can learn appearance and motion from different sources. Note that DyNCA does not simply copy the target dynamics from the video, but learns the style of the given motion and naturally adapts it to the target appearance, as shown by the optical-flow images. See our real-time interactive demo \href{https://dynca.github.io}{https://dynca.github.io}}
\label{fig:video-result}
\vspace{-13pt}
\end{table*}

\vspace{-3pt}
We conduct experiments on the critical parameters of DyNCA, including $N_C, N_H, N_W, N_{FC}$, where $N_C, N_H, N_W$ are the size of the DyNCA seed state and $N_{FC}$ is the output dimensionality of the first layer of MLP in DyNCA. Concretely, we set $N_C=12, N_{FC}=96$ for \textbf{DyNCA-S} and $N_C=16, N_{FC}=128$ for \textbf{DyNCA-L}. Although DyNCA can adapt to arbitrary seed sizes after training, the seed resolution used during training affects the texture details. We experiment with both $128^2$ and $256^2$ spatial resolutions for the seed. We use \textit{Positional Encoding} for all of the DyNCA configurations, and enable \textit{Multi-Scale Perception} only when training with the $256 \times 256$ seed size.
Table~\ref{tab:vid-time-compare} summarizes the DyNCA configurations. We perform our experiments on an Nvidia-A100 GPU, and use Adam optimizer with an initial learning rate of 0.001. We refer the reader to the supplementary for further training details. 

% yielding \textbf{DyNCA-128} and \textbf{DyNCA-256}. Therefore, 4 configurations of DyNCA are \textbf{DyNCA-S-128}, \textbf{DyNCA-S-256}, \textbf{DyNCA-L-128}, \textbf{DyNCA-L-256}. 

% Hardware, optimizer, models, etc...
% \input{sections/Experiments/Appr}

% \subsection{Motion from Vector field}
% \label{sec:exp-mvec}

\vspace{-3pt}
\subsection{Dynamic Texture Synthesis}
\vspace{-3pt}
\label{sec:exp-mvid}

% \usepackage{graphicx}

\newcolumntype{d}[1]{D{.}{.}{#1}}

\begin{table}[t]
\resizebox{\linewidth}{!}{%
\begin{tabular}{cc|c|c|c}
\toprule
% \begin{tabular}[c]{@{}c@{}}\textbf{Sketch Synthesis} \\ \textbf{Algorithms}\end{tabular} & 
\begin{tabular}[c]{@{}c@{}}{Method}\end{tabular} &
\begin{tabular}[c]{@{}c@{}}{Res.}\end{tabular} &
\begin{tabular}[c]{@{}c@{}}{Training Time (s)}\end{tabular} &
\begin{tabular}[c]{@{}c@{}}{Synthesis Time (s)}\end{tabular}  &
\begin{tabular}[c]{@{}c@{}}{\# Parameters}\end{tabular} 
\\
% \begin{tabular}[c]{@{}c@{}}{\textbf{G}}\end{tabular} & \\

\midrule 

A \cite{two_stream} & $256^2$  & \#frames $\times$ 500 & $5.0 \times 10^{2}$ & \#frames $\times$ 0.2M \\

B \cite{xie2017generativeconvnet} &$224^2$ & \#frames $\times$ 400 & $4.0 \times 10^{2}$ & 81M \\

C \cite{xie2017generativeconvnet} & $100^2$ & \#frames $\times$ 8.5 & $8.5 \times 10^{0}$ & 2.8M\\

\midrule
\textbf{DyNCA-S} & $128^2$ & 2320 & $3.3 \times 10^{-2}$ & 0.006M \\
\textbf{DyNCA-S} & $256^2$ & 3980 & $5.7 \times 10^{-2}$ & 0.006M \\
\textbf{DyNCA-L} & $128^2$ & 2370 & $3.5 \times 10^{-2}$ & 0.010M \\
\textbf{DyNCA-L} & $256^2$ & 4380 & $5.7 \times 10^{-2}$ & 0.010M \\
\bottomrule
\end{tabular}
}
\vspace{-7pt}
\caption{Comparison of training time, synthesis time per frame, and number of trainable parameters of different DyTS methods.  (A) Tesfaldet et al. \cite{two_stream}; (B) Xie et al. \cite{xie2017generativeconvnet} FC config; (C) Xie et al. \cite{xie2017generativeconvnet} ST config.  For (B) and (C), training and synthesis happen simultaneously, and the video is synthesized once the training is finished. All methods are evaluated on a single A100 40GB GPU.
}
\label{tab:vid-time-compare}
\vspace{-15pt}
\end{table}

% Although STGConvNet introduced by Xie et al. can generate a video sample only by langevin dynamics after training, it is not capable of generating new frames that do not exist in the original video.

% Comparison of GAN-based image editing algorithms by their characteristics. \fbox{\textbf{A}} Works on any Dataset, \fbox{\textbf{B}} No test-time optimization, \fbox{\textbf{C}} Works on any GAN architecture, \fbox{\textbf{D}} Can perform the edit using a single image, \fbox{\textbf{E}} Allows global semantic editing, \fbox{\textbf{F}} Allows localized semantic editing, \fbox{\textbf{G}} Allows editing any object in the image

\newcommand{\imgmotvec}[1]{\includegraphics[height=200pt]{figures/Experiments/MotionVec/#1}}
\newcolumntype{L}{>{\centering\arraybackslash} m{18pt} } 
\begin{table}[]
\resizebox{\linewidth}{!}{
\begin{tabular}{cccccccccc}

\imgmotvec{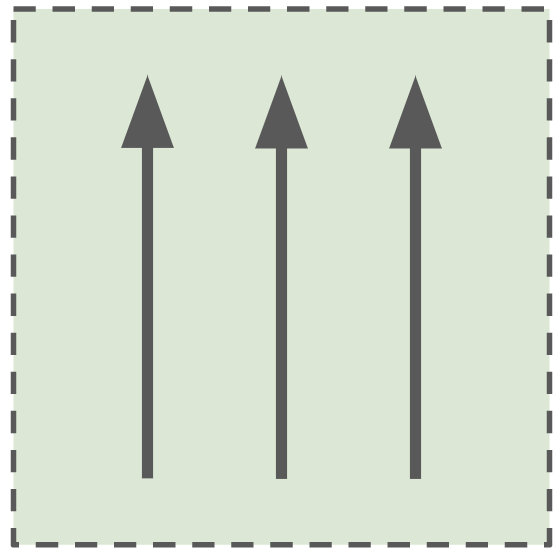} & \imgmotvec{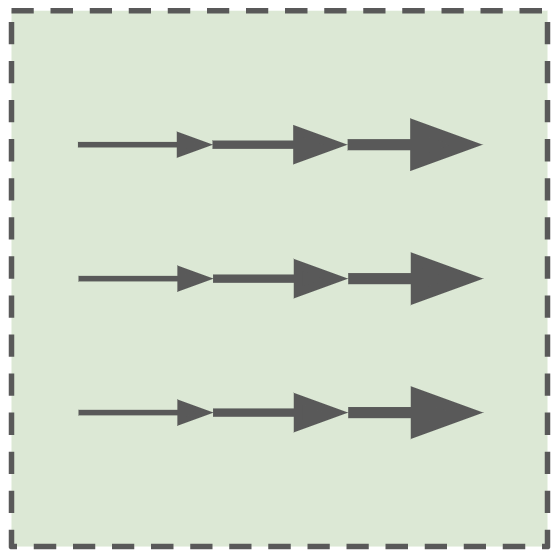} &
\imgmotvec{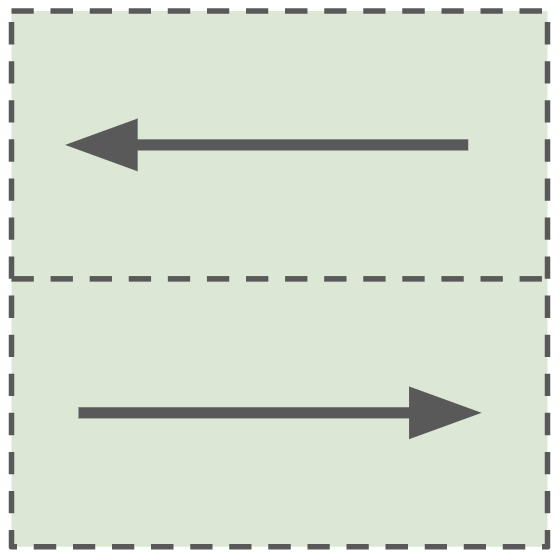} &
\imgmotvec{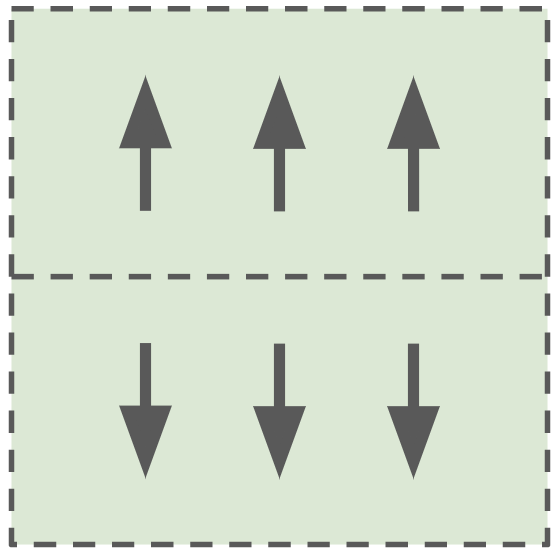} &
\imgmotvec{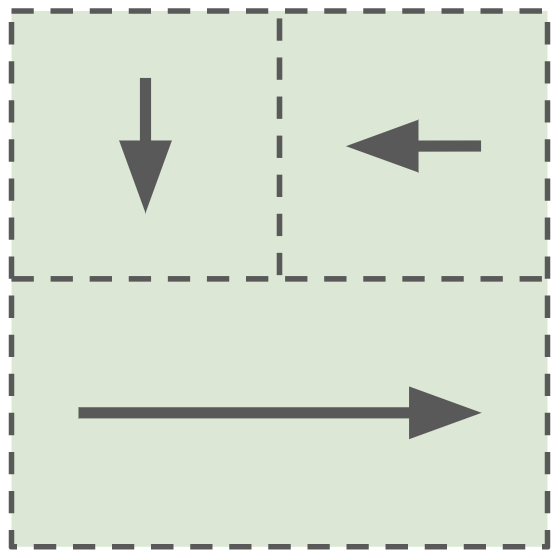} &
\imgmotvec{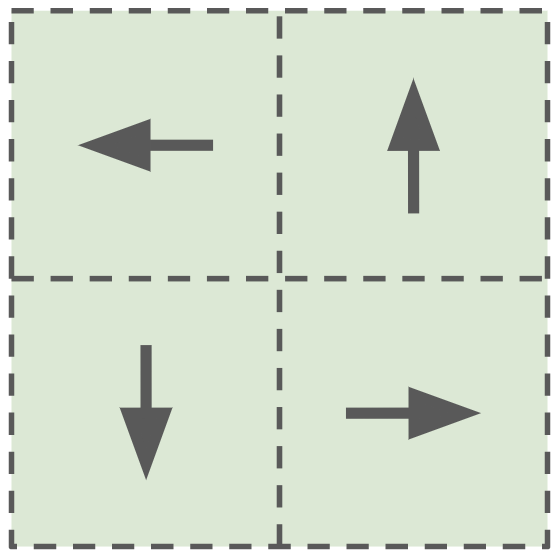} &
\imgmotvec{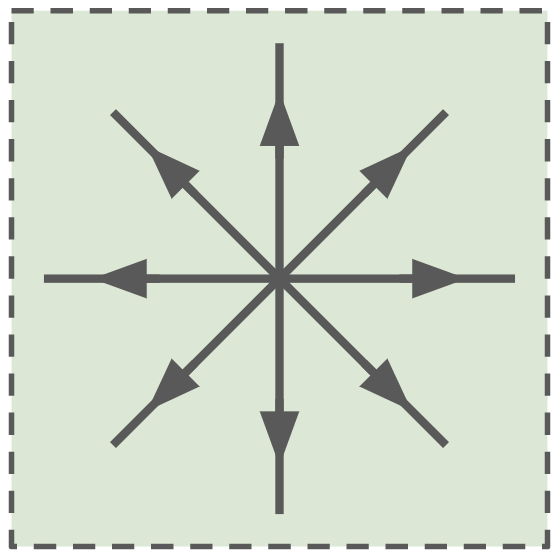} &
\imgmotvec{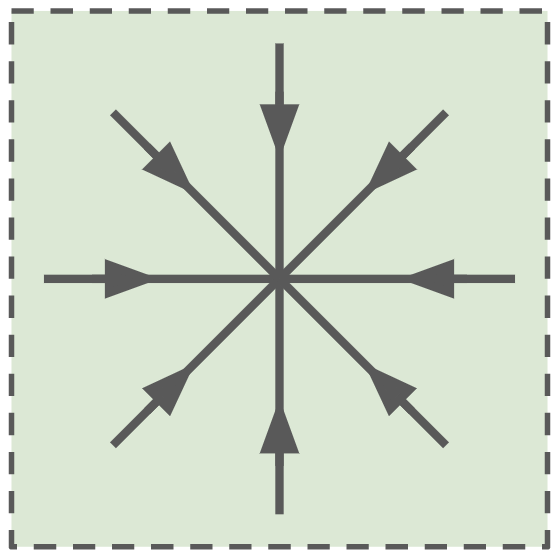} &
\imgmotvec{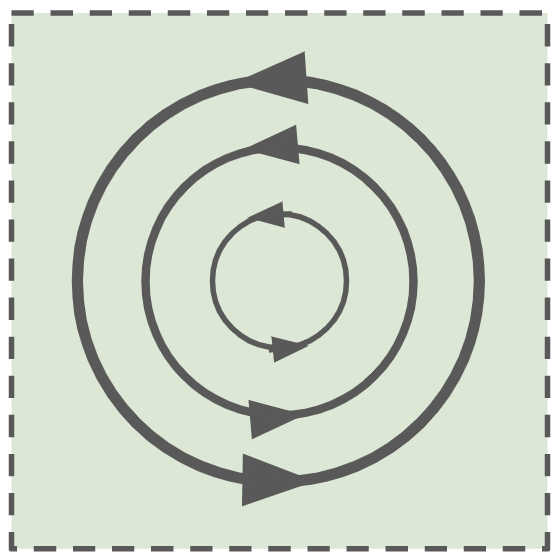} &
\imgmotvec{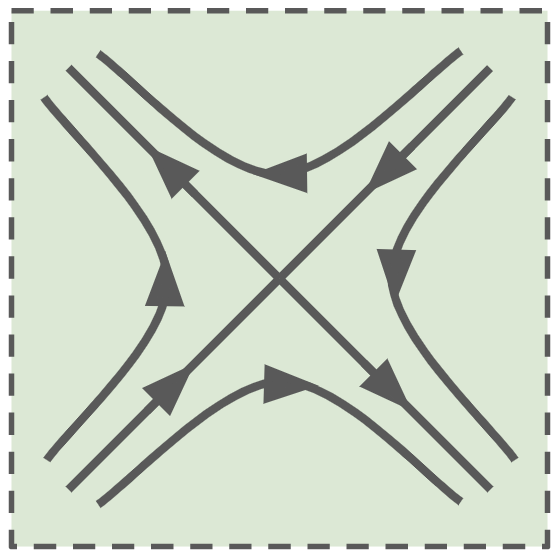}
\\

\imgmotvec{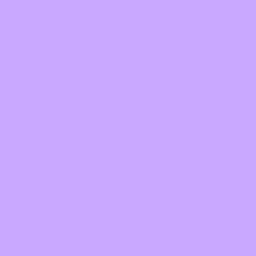} &
\imgmotvec{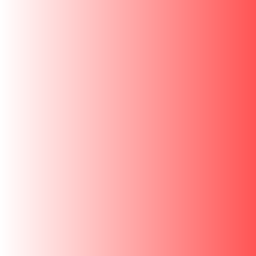} &
\imgmotvec{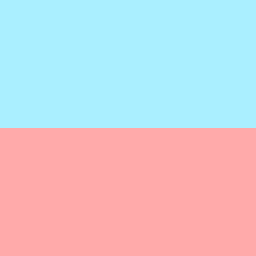} &
\imgmotvec{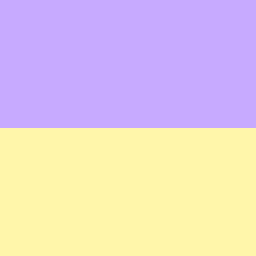} &
\imgmotvec{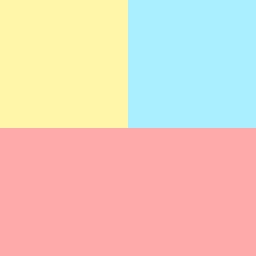} &
\imgmotvec{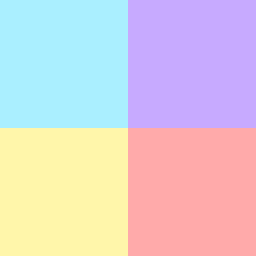} &
\imgmotvec{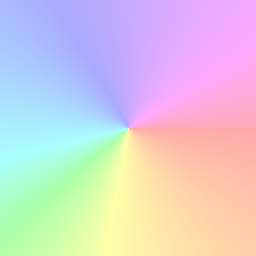} &
\imgmotvec{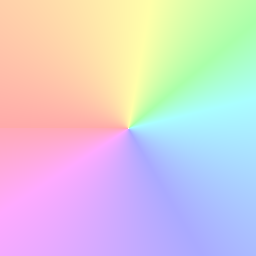} &
\imgmotvec{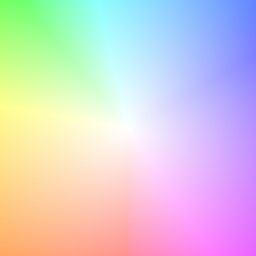} &
\imgmotvec{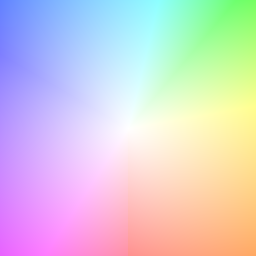} \\

\imgmotvec{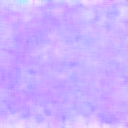} &
\imgmotvec{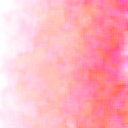} &
\imgmotvec{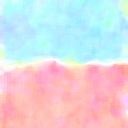} &
\imgmotvec{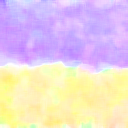} &
\imgmotvec{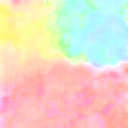} &
\imgmotvec{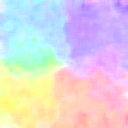} &
\imgmotvec{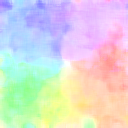} &
\imgmotvec{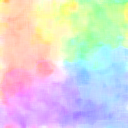} &
\imgmotvec{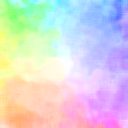} &
\imgmotvec{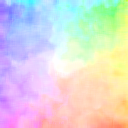} 
\end{tabular}
}
\vspace{-8pt}
\captionof{figure}{Results of DyNCA trained on various target vector fields as the motion target. \textbf{First and second row}: Symbolic and colored optical-flow representations of the target vector fields. \textbf{Third row}:
Snapshot of the optical-flow estimations of the videos synthesized by DyNCA-S-$128^2$. The target appearance used for training the model is fibrous\_0145 texture from the DTD dataset \cite{dtd}.}
\label{fig:vec_field_results}
\vspace{-15pt}
\end{table}

We present the results of synthesizing dynamic textures using DyNCA, where the target motion is either a vector field or a video, and the target appearance is a static image.

\textbf{Motion from Vector Field:}
We manually design 12 target vector fields, and for each target motion, we train both DyNCA-S and DyNCA-L configurations on 45 different target appearances. We provide the resulting 1080 trained models in our real-time interactive demo. We set  seed size to $128\times128$ and $T=24$, assuming that 24 steps of DyNCA update equals one frame of the synthesized video. 
Figure~\ref{fig:vec_field_results} shows the optical flow of the DyNCA-S synthesized videos and the corresponding target vector fields used for training.

\textbf{Motion from Video}. We train DyNCA to match the target motion from videos. For the target appearance image, we use one of the video frames or a stylistic image, to perform dynamic texture synthesis and dynamic style transfer, respectively. We train both DyNCA-S and DyNCA-L on 59 target dynamic texture videos provided in \cite{two_stream}, setting seed size to $256 \times 256$ and $T=64$. Figure \ref{fig:video-result} provides some visual results. More results are provided in our demo and supplementary. In Table \ref{tab:vid-time-compare}, we compare our DyNCA and the previous SOTA models \cite{two_stream, xie2017generativeconvnet} in terms of the computational costs, namely performance and the number of parameters. Note that our DyNCA model is orders of magnitude more efficient in both training and synthesis time as well as the number of parameters. We refer the reader to the supplementary material for the detailed experimental setup.

% Indeed, DyNCA can synthesize realistic videos in real-time.

% \yx{Only Idea now}
% Parameter claim and show figure.  (b)(c) for dynamic texture synthesis and (d) for dynamic style transfer in table \ref{tab:main-res}. \\

% Training time and synthesize time comparison. \\

% Adaptive loss weight settings. First train for 1000 epochs with a fixed weight and reset NCA pools and NCA itself. Obtain the median of the motion loss. Multiply the median loss by a constant plus a bias. The reason for reset all things is because the fixed weight during the first 1000 epochs of training might cause DyNCA getting stuck in some inescapable local minimum and generating wrong motion or appearance. 

% \input{figures/Experiments/MotionVidDST}

\vspace{-3pt}
\subsection{Real-time Video Editing}
\vspace{-3pt}
DyNCA can also perform real-time video editing by utilizing the post-training NCA controls introduced in \cite{niklasson2021self-sothtml}. These edits include direction control, speed control, a brush tool, and local coordinate transformations. We refer the reader to our online demo at \href{https://dynca.github.io}{https://dynca.github.io} for real-time and interactive visualization and experimentation. 

\noindent \textbf{Direction Control:} By rotating the Sobel convolution kernels $\nabla_\textup{x}$ and $\nabla_\textup{y}$ in the perception stage, we can control the direction of the motion in the synthesized video. This is done by replacing $\nabla_\textup{x}$ and $\nabla_\textup{y}$ with $\nabla_\textup{u}$ and $\nabla_\textup{v}$, where $\binom{\textup{u}}{\textup{v}} = \textup{R}_{\theta} \binom{\textup{x}}{\textup{y}}$ and $\textup{R}_{\theta}$ is the rotation matrix with angle $\theta$. We also rotate $\textup{P}_{\textup{x}\textup{y}}$ by angle $\theta$ so that the position-dependent part of the motion magnitude also rotates accordingly.

\noindent \textbf{Speed Control:} We control the speed of the synthesized video by increasing/decreasing $T$, which determines how many DyNCA steps are used to synthesize one video frame.

\noindent \textbf{Brush Tool:} The brush tool allows the users to delete pixels from the video. Since our DyNCA model exhibits the self-organization property of the original NCA \cite{niklasson2021self-sothtml}, the model can reorganize itself and continue synthesizing realistic videos by naturally filling the deleted pixels. 

\noindent \textbf{Local Coordinate Transformation:} DyNCA can create complex motion transformations by allowing each cell to use a different rotation angle $\theta(i,j)$ for transforming its Sobel convolution kernels. For example, setting the rotation angle for the cell at location $(i, j)$ to $arctan \left ( \frac{i - \frac{W}{2}}{j - \frac{H}{2}} \right )$ will transform a rightward motion into a circular motion.

\vspace{-3pt}
\subsection{User Study}
\begin{figure}[htp]

\subfloat{%
  \includegraphics[width=\linewidth]{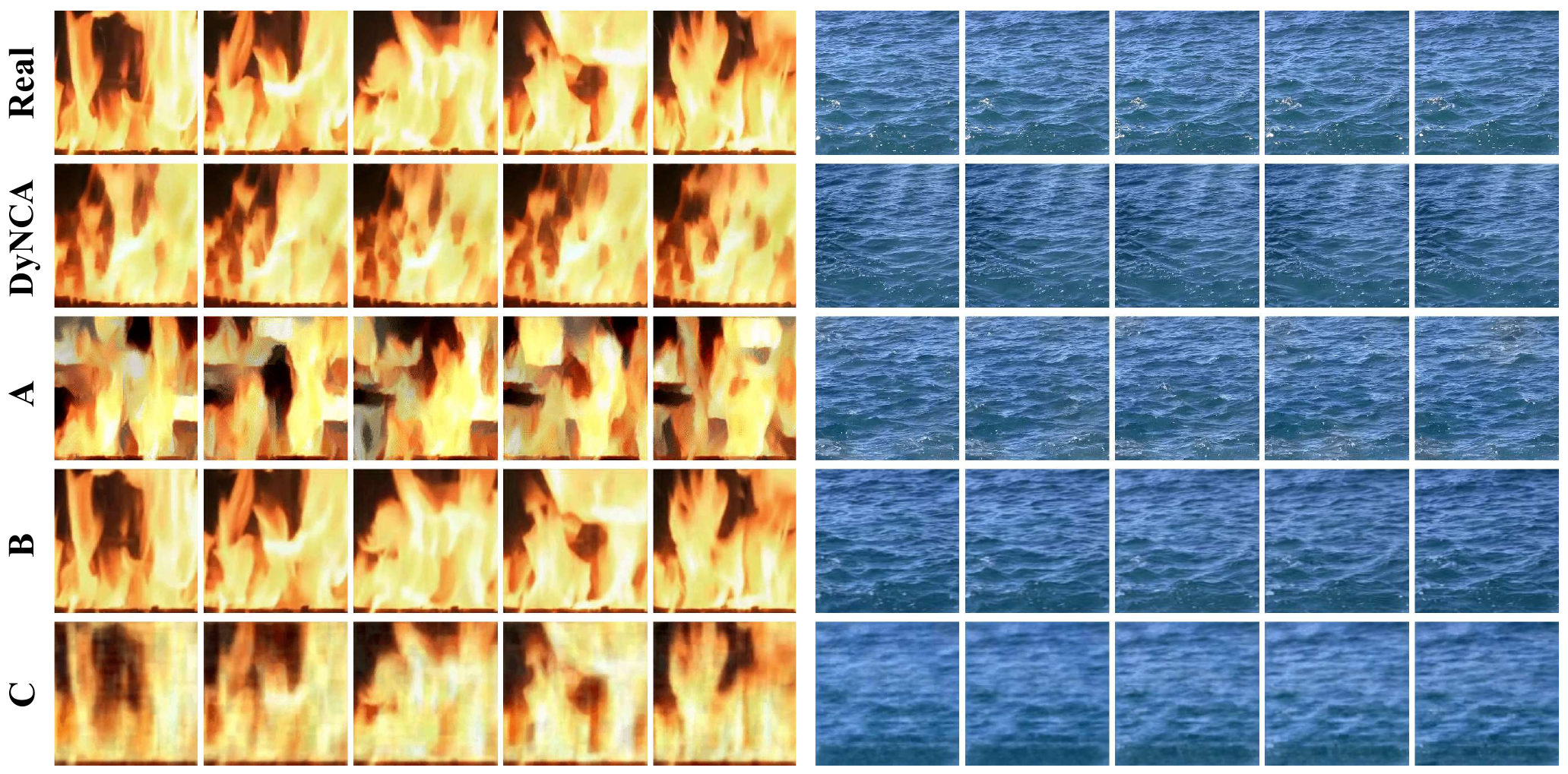}  
}
\vspace{1pt}

\subfloat{
\resizebox{\linewidth}{!}{
\begin{tabular}{cc||cccc}
\toprule
{} &                           Real &                          DyNCA &            A \cite{two_stream} & B \cite{xie2017generativeconvnet} & C \cite{xie2017generativeconvnet} \\
\midrule
Real                              &                             N/A &  \cellcolor[HTML]{FA9857}{27\%} &  \cellcolor[HTML]{F99355}{26\%} &     \cellcolor[HTML]{F8864F}{24\%} &      \cellcolor[HTML]{CA2427}{8\%} \\
DyNCA                             &  \cellcolor[HTML]{91D068}{73\%} &                             N/A &  \cellcolor[HTML]{FEDE89}{40\%} &     \cellcolor[HTML]{FFF1A8}{46\%} &     \cellcolor[HTML]{F46D43}{20\%} \\
A \cite{two_stream}               &  \cellcolor[HTML]{8CCD67}{74\%} &  \cellcolor[HTML]{D7EE8A}{60\%} &                             N/A &     \cellcolor[HTML]{F7FCB4}{52\%} &     \cellcolor[HTML]{F98E52}{25\%} \\
B \cite{xie2017generativeconvnet} &  \cellcolor[HTML]{7FC866}{76\%} &  \cellcolor[HTML]{EEF8A8}{54\%} &  \cellcolor[HTML]{FFF8B4}{48\%} &                                N/A &     \cellcolor[HTML]{E44C34}{15\%} \\
C \cite{xie2017generativeconvnet} &  \cellcolor[HTML]{138C4A}{92\%} &  \cellcolor[HTML]{66BD63}{80\%} &  \cellcolor[HTML]{87CB67}{75\%} &     \cellcolor[HTML]{3CA959}{85\%} &                                N/A \\
\bottomrule
\end{tabular}
}
}
\vspace{-6pt}
\captionof{table}{\textbf{Top: } Comparison of real and synthesized video frames from different methods: (A) Tesfaldet et al. \cite{two_stream}; (B) Xie et al. \cite{xie2017generativeconvnet} FC config; (C) Xie et al. \cite{xie2017generativeconvnet} ST config. \textbf{Bottom: } Pair-wise comparison results from our user study. The participants see two videos one after another in random order and are asked to choose the video that appears more realistic. Our DyNCA achieves an on-par ``fooling" rate when compared with real videos. Yet, our synthesized videos look more realistic when compared to existing DyTS methods (60\%, 54\%, 80\%). All error margins are $\leq 2\%$.%\sabine{What does that mean? It either is or is not 2 percent?}.
}
\label{fig:user-study}
\vspace{-11pt}
\end{figure}

\vspace{-3pt}
\label{sec:exp-userstudy}
Although it is tempting to use a metric, e.g. the Gram matrix difference \cite{gatys2015texture}, to compare the quality of the videos synthesized by different methods, the variety of training objectives used in the existing methods makes such comparisons biased and unfair. Therefore, we conduct a similar user study as Tesfaldet et al. \cite{two_stream} to quantitatively evaluate and compare the realism of the videos.
% As there is no metric for measuring the realism of videos, we conduct a similar user study as Tesfaldet et al. \cite{two_stream} to quantitatively evaluate and compare the realism of the videos synthesized by different methods.
We show a pair of videos to the participants and ask them to choose the video that appears more realistic. We compare the videos synthesized by (DyNCA, \cite{two_stream}, \cite{xie2017generativeconvnet}) with each other and also with real videos using the same 59 dynamic texture videos as Tesfaldet et al. \cite{two_stream}. Table~\ref{fig:user-study} shows the results of our user study. Each entry shows the percentage of times that the video from the corresponding column was chosen over the video from the corresponding row. Our method outperforms the other DyTS method in  realism.
We refer the reader to our supplementary for more details on the user study.

% Note that our DyNCA method is on par or even outperforms the other methods on realism.  details. %\sabine{you have to give a lot more details in the suppl. than you had here. But you can gain space because you do not need to talk about number of observers.} 

% % Please add the following required packages to your document preamble:
% % \usepackage{graphicx}

% \begin{table}[t]
% \resizebox{\linewidth}{!}{%
% \begin{tabular}{c|c|c}
% \toprule
% % \begin{tabular}[c]{@{}c@{}}\textbf{Sketch Synthesis} \\ \textbf{Algorithms}\end{tabular} & 
% \begin{tabular}[c]{@{}c@{}}{Method}\end{tabular} &
% \begin{tabular}[c]{@{}c@{}}{Appearance Similarity}\end{tabular} &
% \begin{tabular}[c]{@{}c@{}}{Motion Similarity}\end{tabular} 
% \\
% % \begin{tabular}[c]{@{}c@{}}{\textbf{G}}\end{tabular} & \\

% \midrule 

% Tesfaldet et al. \cite{two_stream} & 4.0 & 4.0 \\

% Xie et al. \cite{xie2017generativeconvnet}  & 4.0 & 4.0 \\

% \midrule
% \textbf{DyNCA-L-256} & 6.0 & 6.0 \\
% \bottomrule
% \end{tabular}
% }
% % \vspace{-5pt}
% \caption{Comparison of appearance and motion similarities judged by humans. 
% }
% \label{tab:vid-user-study}
% \end{table}

% % Comparison of GAN-based image editing algorithms by their characteristics. \fbox{\textbf{A}} Works on any Dataset, \fbox{\textbf{B}} No test-time optimization, \fbox{\textbf{C}} Works on any GAN architecture, \fbox{\textbf{D}} Can perform the edit using a single image, \fbox{\textbf{E}} Allows global semantic editing, \fbox{\textbf{F}} Allows localized semantic editing, \fbox{\textbf{G}} Allows editing any object in the image

%We limit the exposure time to the original length of the video to avoid looping artefacts. 
% Sentinel videos
% Warmup videos

\vspace{-5pt}
\subsection{Ablation Study}
\label{sec:exp-ablation}

% \begin{itemize}
\vspace{-5pt}
\textbf{Positional Encoding:} We ablate the Cartesian Positional Encoding (CPE) and train DyNCA-S-$128^2$ with different padding strategies for comparison. In Table~\ref{tab:pos_emb_ablation}, we report the average of $\mathcal{L}_{dir}$ and $\mathcal{L}_{norm}$ losses on 4 target vector fields (\textit{Diverge}, \textit{Converge}, \textit{Circular}, and \textit{Hyperbolic} shown in the last 4 columns of Figure~\ref{fig:vec_field_results}) and 10 different target appearances. The results demonstrate that CPE is a necessary component for DyNCA to learn the motion from a structured target vector field, in which the motion direction and magnitude are position-dependent. We show the importance of CPE for video motion fitting in the supplementary.

%  For video motion, we perform a  quantitative evaluation to show the effectiveness of CPE on video motion learning. We record the appearance loss and the video motion loss during video synthesis after training across all texture images in DynTex. The loss values show the effectiveness of CPE in helping DyNCA fit the target appearance and motion. Values are in the supplementary.

\textbf{Multi-Scale Perception:}. MSP is key to enable training DyNCA with larger seed sizes. We train DyNCA-L-$256^2$ both with and without multi-scale perception and evaluate the results qualitatively and quantitatively. Figure~\ref{fig:multiscale-abl} shows the corresponding synthesized video frames, qualitatively demonstrating that single-scale perception causes artifacts to appear in the synthesized frames. We also perform a quantitative evaluation in the supplementary, showing that training with multi-scale perception improves both the appearance $\mathcal{L}_{appr}$ and the video motion $\mathcal{L}_{mvid}$ losses.

    % \item The update steps of DyNCA to match one-frame motion. 16 32 cannot work, 48 is OK but contain artefacts (fireplace2 256), 64 is maximum memory usage for 256.  
    % % \item Adaptive loss weight (Vec-field and video motion)
    % % Compare with fixed loss setting. fireplace-2 will be fixed.
    % \item Other optic flow network (RAFT, FlowNETS)
    % Wrong estimation of motion on texture (e.g., birds,gif)

% \end{itemize}

\vspace{-6pt}
\section{Limitations}
\vspace{-2pt}

DyNCA has certain limitations. In the case of vector field motion, it cannot generate a correct motion when the target appearance and the target motion are not compatible. For example, DyNCA fails to move a 45-degree oblique-line pattern in a circular manner. For video motion, we find it difficult to automatically set the motion loss weight $\lambda$. 
%We observe 5 diverging, and 2 low-quality cases due to too-large or too-small motion loss weights. 
Moreover, DyNCA cannot generate diverse motion when the target videos are violating the underlying assumption of dynamic textures, i.e. when they are not temporally homogeneous. In these cases, DyNCA suffers from overfitting to the dominant direction of the motion in the target videos
%, and synthesizes videos that only contain the dominant motion. 
We refer to our supplementary material for more details.

\vspace{-6pt}
\section{Conclusion}
\vspace{-2pt}
\label{sec:conclusion}

\begin{table}[t!]
\resizebox{\linewidth}{!}{
\begin{tabular}{cc|cccc}
\toprule
\textbf{Seed Size}                & \textbf{Loss}      & Circular & Zero & Replicate & CPE \\
\midrule

\multirow{2}{*}{$128\times 128$} & $\mathcal{L}_{dir}$  & 0.989    & 0.162 & 0.318     & \textbf{0.062}  \\
                         & $\mathcal{L}_{norm}$ & 0.640    & 0.296 & 0.364     & \textbf{0.235}  \\
\midrule
\multirow{2}{*}{$256\times 256$} & $\mathcal{L}_{dir}$  & 0.993    & 0.411 & 0.478     & \textbf{0.054}  \\
                         & $\mathcal{L}_{norm}$ & 0.678    & 0.331 & 0.397     & \textbf{0.218} \\
\bottomrule
\end{tabular}
}
\vspace{-7pt}
\caption{Comparison of motion loss for  DyNCA trained with different padding strategies. Using Cartesian Positional Encoding (CPE) improves the motion fidelity, and allows the DyNCA to synthesize the correct motion regardless of the seed size. %These results are evaluated by averaging the losses over 40 trainings (10 different target appearances, and 4 different target motions (circular, hyperbolic, converge, and diverge).
}
\label{tab:pos_emb_ablation}
\vspace{-8pt}
\end{table}    

\newcommand{\imgmvidabl}[1]{\includegraphics[height=70pt]{figures/Experiments/MultiscaleAbl/#1}}

\newcolumntype{S}{>{\centering\arraybackslash} m{60pt} } 
\newcolumntype{Q}{>{\centering\arraybackslash} m{2pt} } 
\begin{table}[t]
\resizebox{\linewidth}{!}{
\begin{tabular}{m{2pt}SSSS}
    
% \toprule
    % \textbf{Target} &
    % \textbf{Single-scale} &
    % \textbf{Multi-scale} \\
    
    % \midrule
    
{\rotatebox{90}{\parbox{2.5cm}{\hspace{19pt} \textbf{Target}}}} &
\imgmvidabl{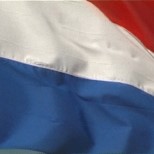}                                & \imgmvidabl{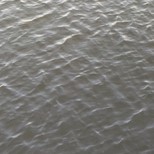}       & 
 \imgmvidabl{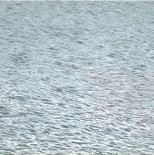}  &      
 \imgmvidabl{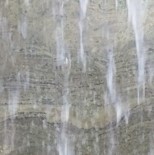} \\

% \midrule  

{\rotatebox{90}{\parbox{2.5cm}{\hspace{8pt} \textbf{Single-scale}}}} &
                          \imgmvidabl{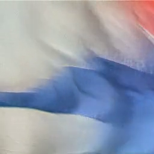}      & \imgmvidabl{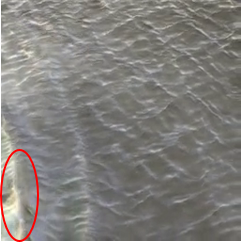}      &     \imgmvidabl{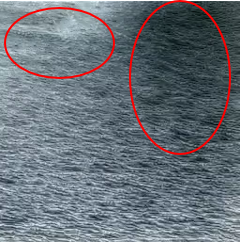}   &    \imgmvidabl{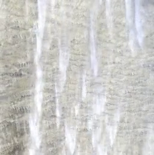}                \\
                            {\rotatebox{90}{\parbox{2.5cm}{\hspace{8pt} \textbf{Multi-scale}}}} &
                            \imgmvidabl{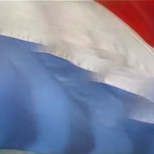}     & \imgmvidabl{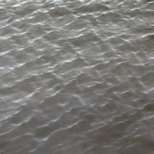}      &     \imgmvidabl{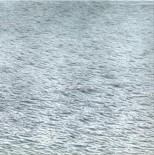} &         
                            \imgmvidabl{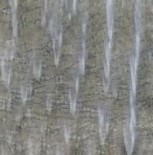}\\

\end{tabular}
}
\vspace{-10pt}
\captionof{figure}{Comparison between training DyNCA with and without multi-scale perception. The first row shows the target appearance texture. The results without multi-scale perception (second row) contain artifacts and are of lower quality.}
\label{fig:multiscale-abl}
\vspace{-10pt}
\end{table}

% \newcommand{\imgmvidsmall}[1]{\includegraphics[height=55pt]{figures/Experiments/MotionVid/#1}}

% \begin{table}[]
% \begin{tabular}{c|c|c}
% % \hline
% \textbf{Target Appearance} & Without MP & With MP \\  \hline

% \imgmvidsmall{water_3.png}&\imgmvidsmall{water_3.png}&\imgmvidsmall{water_3.png}\\ 
% \imgmvidsmall{water_3.png}&\imgmvidsmall{water_3.png}&\imgmvidsmall{water_3.png}\\ 
% \imgmvidsmall{water_3.png}&\imgmvidsmall{water_3.png}&\imgmvidsmall{water_3.png}\\ 
% \end{tabular}
% \caption{111}
% \label{tab:multiscale-abl}
% \end{table}

We propose DyNCA, a model that can, in real time, synthesize dynamic texture videos with arbitrary frame size and infinite length. Exploiting multi-scale perception and positional encoding, the cells in DyNCA can readily perform long-range communication and obtain global information. This ensures improved performance both in terms of visual quality and computational expressivity as compared to the vanilla NCA model, as we show with both qualitative and quantitative experiments. DyNCA can learn motion either from a hand-crafted vector field or a video, thus allowing for broader synthesis options. DyNCA produces more realistic video textures than the current DyTS methods, as demonstrated through a user study. DyNCA is also $2 \sim 4$ orders of magnitude faster than the SOTA methods in synthesis time and has much fewer trainable parameters, thus facilitating real-world deployment. Lastly, DyNCA allows for several real-time and interactive video control tools that let the users control DyNCA without re-training.

\noindent \small{\textbf{Acknowledgement.} This work was supported in part by the Swiss National Science Foundation via the Sinergia grant CRSII5-180359.}

\clearpage
%%%%%%%%% REFERENCES

{\small
\bibliographystyle{ieee_fullname}
\bibliography{main}

\begin{thebibliography}{10}\itemsep=-1pt

\bibitem{adelson1984pyramid}
Edward~H Adelson, Charles~H Anderson, James~R Bergen, Peter~J Burt, and Joan~M
  Ogden.
\newblock Pyramid methods in image processing.
\newblock {\em RCA engineer}, 29(6):33--41, 1984.

\bibitem{flow_visualization}
Simon Baker, Daniel Scharstein, JP Lewis, Stefan Roth, Michael~J Black, and
  Richard Szeliski.
\newblock A database and evaluation methodology for optical flow.
\newblock {\em International Journal of Computer Vision}, 92(1):1--31, 2011.

\bibitem{burt1987laplacian}
Peter~J Burt and Edward~H Adelson.
\newblock The laplacian pyramid as a compact image code.
\newblock In {\em Readings in Computer Vision}, pages 671--679. Elsevier, 1987.

\bibitem{dtd}
M. Cimpoi, S. Maji, I. Kokkinos, S. Mohamed, , and A. Vedaldi.
\newblock Describing textures in the wild.
\newblock In {\em Proceedings of the {IEEE} Conference on Computer Vision and
  Pattern Recognition ({CVPR})}, 2014.

\bibitem{costantini2007higherordersvd}
Roberto Costantini, Luciano Sbaiz, and Sabine Susstrunk.
\newblock Higher order svd analysis for dynamic texture synthesis.
\newblock {\em IEEE Transactions on Image Processing}, 17(1):42--52, 2007.

\bibitem{doretto2003dynamic-firstds}
Gianfranco Doretto, Alessandro Chiuso, Ying~Nian Wu, and Stefano Soatto.
\newblock Dynamic textures.
\newblock {\em International Journal of Computer Vision}, 51(2):91--109, 2003.

\bibitem{doretto2004spatially}
Gianfranco Doretto, Eagle Jones, and Stefano Soatto.
\newblock Spatially homogeneous dynamic textures.
\newblock In {\em European Conference on Computer Vision}, pages 591--602.
  Springer, 2004.

\bibitem{funke2017synthesising-gatysdynamic}
Christina~M Funke, Leon~A Gatys, Alexander~S Ecker, and Matthias Bethge.
\newblock Synthesising dynamic textures using convolutional neural networks.
\newblock {\em arXiv preprint arXiv:1702.07006}, 2017.

\bibitem{gatys2015texture}
Leon Gatys, Alexander~S Ecker, and Matthias Bethge.
\newblock Texture synthesis using convolutional neural networks.
\newblock {\em Advances in Neural Information Processing Systems}, 28, 2015.

\bibitem{ghadekar2014nonlinear}
PP Ghadekar and NB Chopade.
\newblock Nonlinear dynamic texture analysis and synthesis model.
\newblock {\em International Journal on Recent Trends in Engineering \&
  Technology}, 11(1):475, 2014.

\bibitem{gilpin2019cellular}
William Gilpin.
\newblock Cellular automata as convolutional neural networks.
\newblock {\em Physical Review E}, 100(3):032402, 2019.

\bibitem{Holynski_2021_CVPR}
Aleksander Holynski, Brian~L. Curless, Steven~M. Seitz, and Richard Szeliski.
\newblock Animating pictures with eulerian motion fields.
\newblock In {\em Proceedings of the IEEE/CVF Conference on Computer Vision and
  Pattern Recognition (CVPR)}, pages 5810--5819, June 2021.

\bibitem{kolkin2019style-otloss}
Nicholas Kolkin, Jason Salavon, and Gregory Shakhnarovich.
\newblock Style transfer by relaxed optimal transport and self-similarity.
\newblock In {\em Proceedings of the IEEE/CVF Conference on Computer Vision and
  Pattern Recognition}, pages 10051--10060, 2019.

\bibitem{sift}
David~G Lowe.
\newblock Object recognition from local scale-invariant features.
\newblock In {\em Proceedings of the Seventh IEEE International Conference on
  Computer Vision}, volume~2, pages 1150--1157. Ieee, 1999.

\bibitem{multiscale_turing}
Jonathan McCabe.
\newblock Cyclic symmetric multi-scale turing patterns.
\newblock In {\em Proceedings of Bridges 2010: Mathematics, Music, Art,
  Architecture, Culture}, pages 387--390, 2010.

\bibitem{mordvinstev_youtube}
Alexander Mordvinstev.
\newblock Fixing neural ca colors with sliced optimal transport.

\bibitem{mordvintsev2021mu-micronca}
Alexander Mordvintsev and Eyvind Niklasson.
\newblock $ \mu $ nca: Texture generation with ultra-compact neural cellular
  automata.
\newblock {\em arXiv preprint arXiv:2111.13545}, 2021.

\bibitem{mordvintsev2020growing}
Alexander Mordvintsev, Ettore Randazzo, Eyvind Niklasson, and Michael Levin.
\newblock Growing neural cellular automata.
\newblock {\em Distill}, 2020.
\newblock https://distill.pub/2020/growing-ca.

\bibitem{niklasson2021self-sothtml}
Eyvind Niklasson, Alexander Mordvintsev, Ettore Randazzo, and Michael Levin.
\newblock Self-organising textures.
\newblock {\em Distill}, 6(2):e00027--003, 2021.

\bibitem{randazzo2020self-classifying}
Ettore Randazzo, Alexander Mordvintsev, Eyvind Niklasson, Michael Levin, and
  Sam Greydanus.
\newblock Self-classifying mnist digits.
\newblock {\em Distill}, 2020.
\newblock https://distill.pub/2020/selforg/mnist.

\bibitem{schodl2000video}
Arno Sch{\"o}dl, Richard Szeliski, David~H Salesin, and Irfan Essa.
\newblock Video textures.
\newblock In {\em Proceedings of the 27th Annual Conference on Computer
  Graphics and Interactive Techniques}, pages 489--498, 2000.

\bibitem{vgg}
Karen Simonyan and Andrew Zisserman.
\newblock Very deep convolutional networks for large-scale image recognition.
\newblock In Yoshua Bengio and Yann LeCun, editors, {\em 3rd International
  Conference on Learning Representations, {ICLR} 2015, San Diego, CA, USA, May
  7-9, 2015, Conference Track Proceedings}, 2015.

\bibitem{soatto2001dynamic}
Stefano Soatto, Gianfranco Doretto, and Ying~Nian Wu.
\newblock Dynamic textures.
\newblock In {\em Proceedings Eighth IEEE International Conference on Computer
  Vision. ICCV 2001}, volume~2, pages 439--446. IEEE, 2001.

\bibitem{two_stream}
Matthew Tesfaldet, Marcus~A Brubaker, and Konstantinos~G Derpanis.
\newblock Two-stream convolutional networks for dynamic texture synthesis.
\newblock In {\em Proceedings of the IEEE Conference on Computer Vision and
  Pattern Recognition}, pages 6703--6712, 2018.

\bibitem{turing1990chemical}
Alan~Mathison Turing.
\newblock The chemical basis of morphogenesis.
\newblock {\em Bulletin of Mathematical Biology}, 52(1):153--197, 1990.

\bibitem{risser2017stable-gramunstable}
Pierre Wilmot, Eric Risser, and Connelly Barnes.
\newblock Stable and controllable neural texture synthesis and style transfer
  using histogram losses.
\newblock {\em CoRR}, abs/1701.08893, 2017.

\bibitem{xie2017generativeconvnet}
Jianwen Xie, Song-Chun Zhu, and Ying Nian~Wu.
\newblock Synthesizing dynamic patterns by spatial-temporal generative convnet.
\newblock In {\em Proceedings of the IEEE Conference on Computer Vision and
  Pattern Recognition}, pages 7093--7101, 2017.

\bibitem{padding_inductive_bias}
Rui Xu, Xintao Wang, Kai Chen, Bolei Zhou, and Chen~Change Loy.
\newblock Positional encoding as spatial inductive bias in gans.
\newblock In {\em arxiv}, December 2020.

\bibitem{yang2016stationary}
Feng Yang, Gui-Song Xia, Liangpei Zhang, and Xin Huang.
\newblock Stationary dynamic texture synthesis using convolutional neural
  networks.
\newblock In {\em 2016 IEEE 13th International Conference on Signal Processing
  (ICSP)}, pages 1135--1139. IEEE, 2016.

\bibitem{you2016kernel}
Xinge You, Weigang Guo, Shujian Yu, Kan Li, Jos{\'e}~C Pr{\'\i}ncipe, and
  Dacheng Tao.
\newblock Kernel learning for dynamic texture synthesis.
\newblock {\em IEEE Transactions on Image Processing}, 25(10):4782--4795, 2016.

\bibitem{yuan2004synthesizing-clds}
Lu Yuan, Fang Wen, Ce Liu, and Heung-Yeung Shum.
\newblock Synthesizing dynamic texture with closed-loop linear dynamic system.
\newblock In {\em European Conference on Computer Vision}, pages 603--616.
  Springer, 2004.

\bibitem{zhang2021dynamic}
Kaitai Zhang, Bin Wang, Hong-Shuo Chen, Xuejing Lei, Ye Wang, and C-C~Jay Kuo.
\newblock Dynamic texture synthesis by incorporating long-range spatial and
  temporal correlations.
\newblock In {\em 2021 International Symposium on Signals, Circuits and Systems
  (ISSCS)}, pages 1--4. IEEE, 2021.

\end{thebibliography}
}

\end{document}